\documentclass[lettersize,journal]{IEEEtran}

\usepackage{times}
\usepackage{soul}
\usepackage{url}
\usepackage[hidelinks]{hyperref}
\usepackage[utf8]{inputenc}
\usepackage[small]{caption}
\usepackage{graphicx}
\usepackage{amsmath}
\usepackage{amsthm}
\usepackage{booktabs}
\usepackage{algorithmic}
\usepackage[switch]{lineno}
\usepackage{adjustbox}
\usepackage{makecell}

\usepackage{tikz,pgfplots}
\pgfplotsset{compat=1.18}
\usepackage{pgfplotstable}
\usepackage{pgfkeys}
\usepgfplotslibrary{groupplots}
\usetikzlibrary{arrows}
\usetikzlibrary{calc}
\usetikzlibrary{fit}
\usetikzlibrary{pgfplots.groupplots}
\usepackage{caption}
\usepackage{subcaption}
\usepackage{color}
\usepackage{amsmath,amssymb,amsfonts}
\usepackage{textcomp}
\usepackage{xcolor}
\usepackage{multirow}
\usepackage{enumitem}
\usepackage[linesnumbered,ruled,vlined]{algorithm2e}
\usepackage{tabularray}
\UseTblrLibrary{booktabs}
\usepackage{pifont}
\usepackage{orcidlink} 

\SetKwInput{KwInput}{Input}
\SetKwInput{KwOutput}{Output}

\begin{document}

\title{Dynamic Configuration of On-Street Parking Spaces using\\ Multi Agent Reinforcement Learning\\}

\author{
\IEEEauthorblockN{
Oshada Jayasinghe\,\orcidlink{0009-0003-3063-1006}, Farhana Choudhury\,\orcidlink{0000-0001-6529-4220},  Egemen Tanin\,\orcidlink{0000-0001-5014-1969}, and Shanika Karunasekera\,\orcidlink{0000-0001-7080-5064}}\\
    \IEEEauthorblockA{The School of Computing and Information Systems, The University of Melbourne}\\
\IEEEauthorblockA{\{jjjayasi@student., farhana.choudhury@, etanin@, karus@\}unimelb.edu.au}
}

\maketitle

\begin{abstract}
With increased travelling needs more than ever, traffic congestion has become a major concern in most urban areas. Allocating spaces for on-street parking, further hinders traffic flow, by limiting the effective road width available for driving. With the advancement of vehicle-to-infrastructure connectivity technologies, we explore how the impact of on-street parking on traffic congestion could be minimized, by dynamically configuring on-street parking spaces. Towards that end, we formulate dynamic on-street parking space configuration as an optimization problem, and we follow a data driven approach, considering the nature of our problem. Our proposed solution comprises a two-layer multi agent reinforcement learning based framework, which is inherently scalable to large road networks. The lane level agents are responsible for deciding the optimal parking space configuration for each lane, and we introduce a novel Deep Q-learning architecture which effectively utilizes long short term memory networks and graph attention networks to capture the spatio-temporal correlations evident in the given problem. The block level agents control the actions of the lane level agents and maintain a sufficient level of parking around the block. We conduct a set of comprehensive experiments using SUMO, on both synthetic data as well as real-world data from the city of Melbourne. Our experiments show that the proposed framework could reduce the average travel time loss of vehicles significantly, reaching upto 47\%, with a negligible increase in the walking distance for parking.
\end{abstract}


\section{Introduction}

\label{introduction}

With rising population and rapid urbanization in the last few decades, increased urban traffic has imposed significant challenges for the existing transportation infrastructure. Improving traffic flow through solutions that adapt to the current traffic conditions, is a key research frontier with emerging intelligent transportation systems. For instance, reinforcement learning (RL) has been widely used for intelligent traffic signal control \cite{prabu, chumarl, wumarl, WANGmarl, mamarl}, to determine the timing phases of the traffic lights. However, less attention has been given to optimizing the usage of on-street parking spaces, which directly contribute to the city-wide traffic by occupying a portion of drivable road area. The existing studies on parking optimization \cite{milp, taiwan, secon, sun2024, markov}, mostly focus on assigning parking spaces to vehicles, i.e., optimization from the users' perspective. We take initiative to look into this problem from the infrastructure's perspective, by dynamically configuring on-street parking spaces, i.e., restricting or allowing parking at each space, based on the observed environmental conditions. The evolution of vehicle-to-infrastructure connectivity technologies paves the way for such dynamic configuration mechanisms, which were previously not feasible. Vehicles can obtain parking restriction information in real-time through mobile phones or in-car display units, similar to popular map based parking availability applications, such as \cite{com_opendata}.

\begin{figure}[t]
    \centering
    \begin{subfigure}[b]{0.24\columnwidth}
        \centering
        \includegraphics[width=0.9\linewidth]{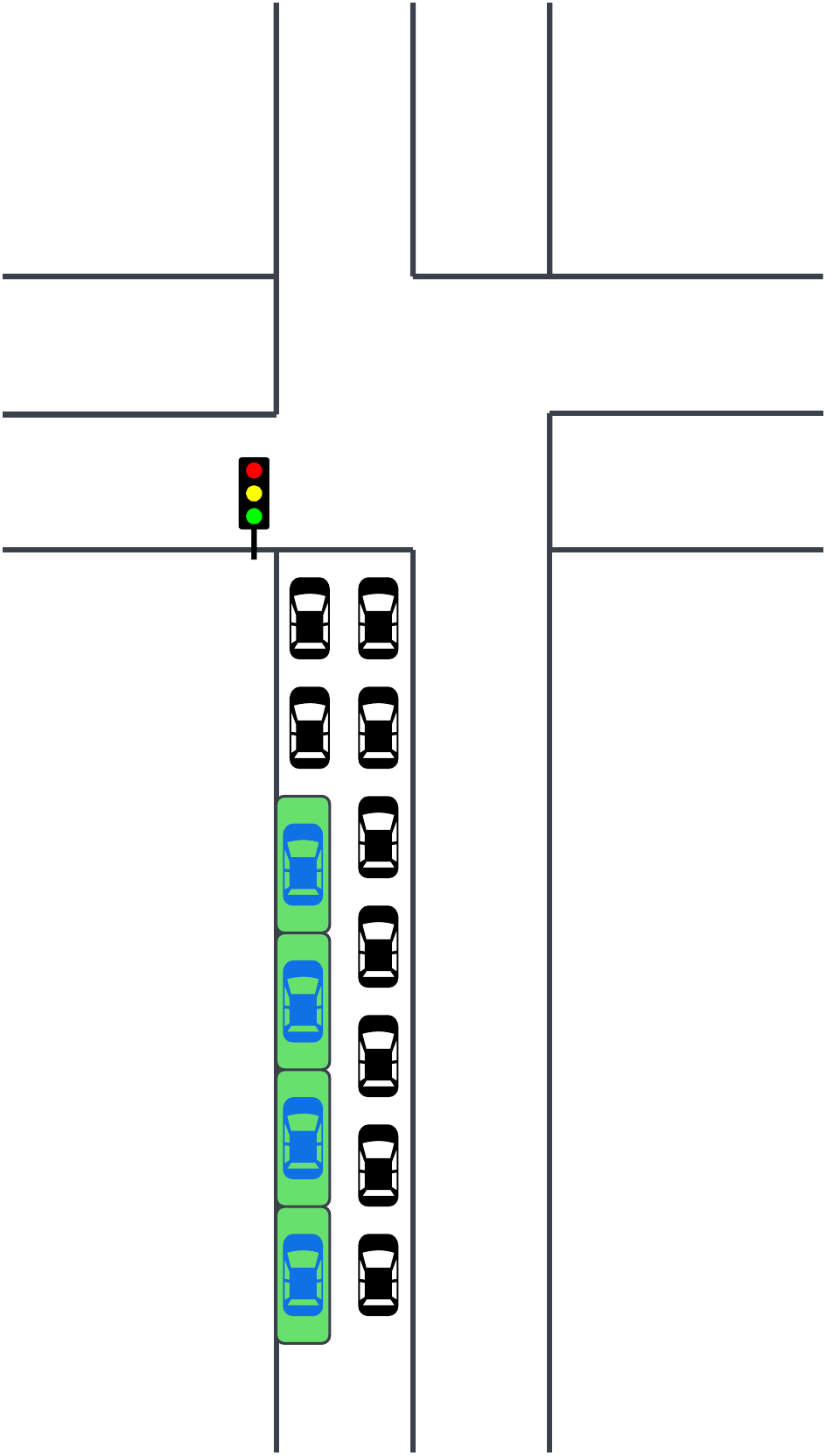}
        \caption{}
        \label{fig:intro_fig1}
    \end{subfigure}
    \begin{subfigure}[b]{0.24\columnwidth}
        \centering
        \includegraphics[width=0.9\linewidth]{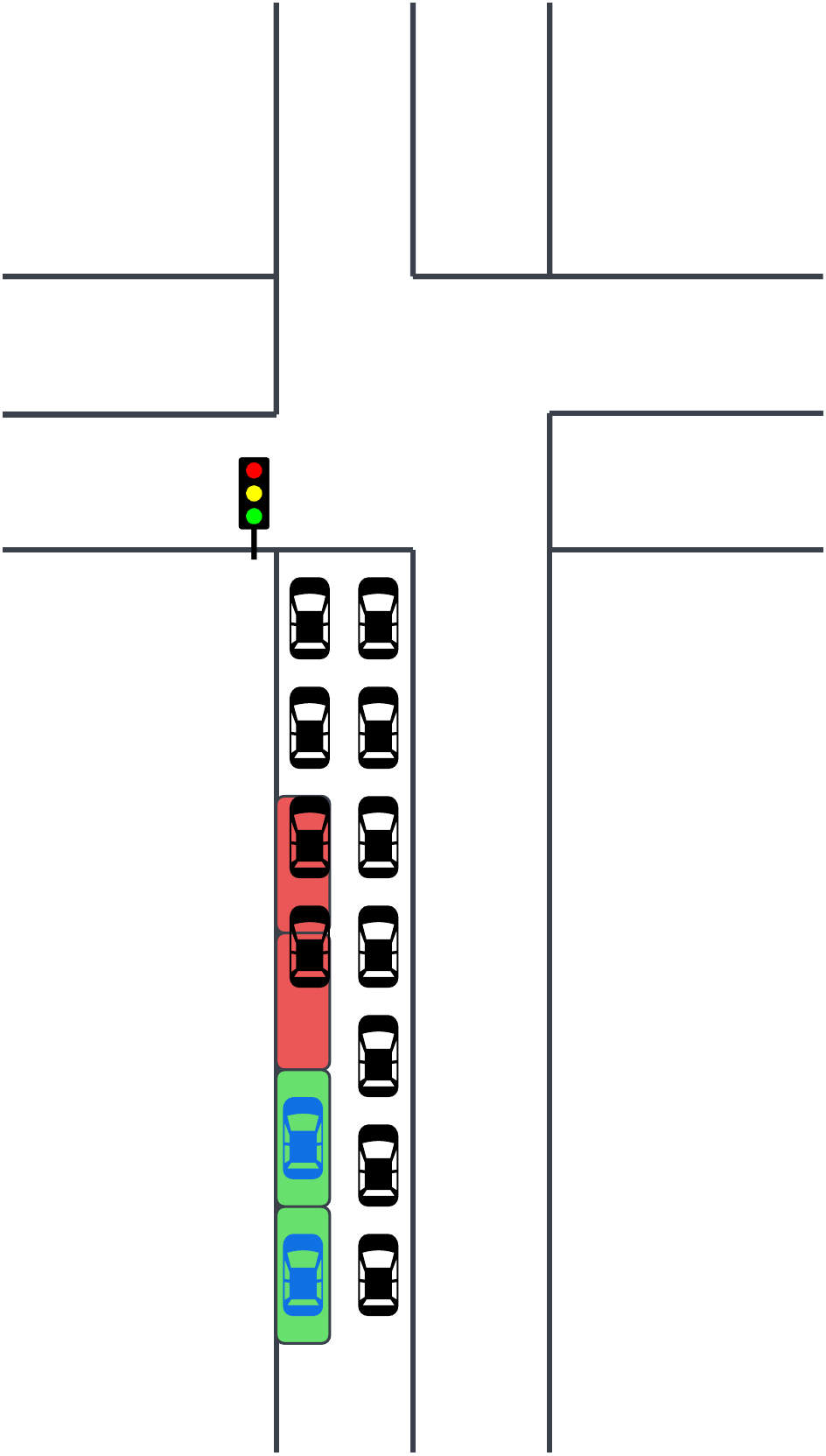}
        \caption{}
        \label{fig:intro_fig2}
    \end{subfigure}
    \begin{subfigure}[b]{0.24\columnwidth}
        \centering
        \includegraphics[width=0.9\linewidth]{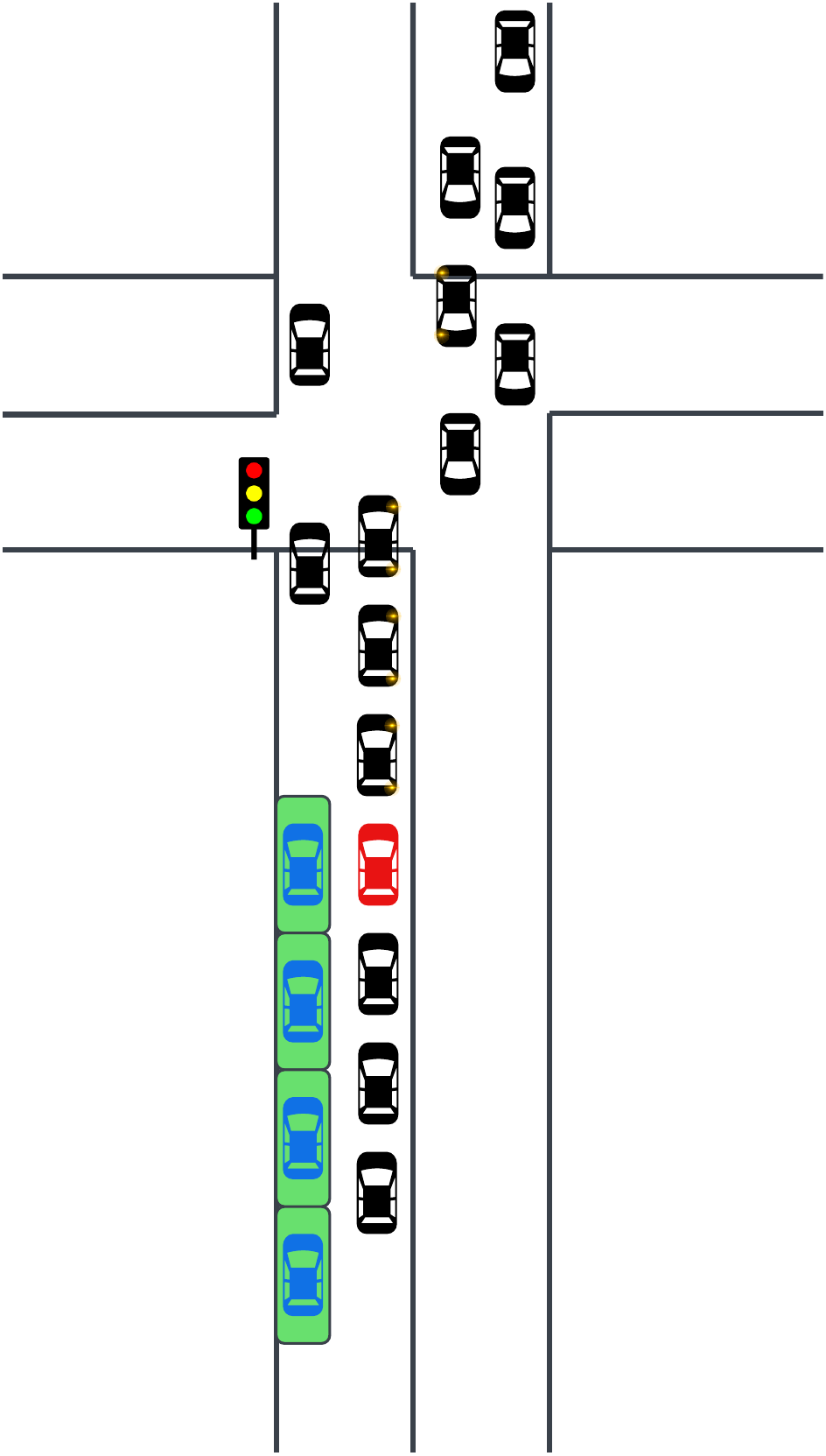}
        \caption{}
        \label{fig:intro_fig3}
    \end{subfigure}
    \begin{subfigure}[b]{0.24\columnwidth}
        \centering
        \includegraphics[width=0.9\linewidth]{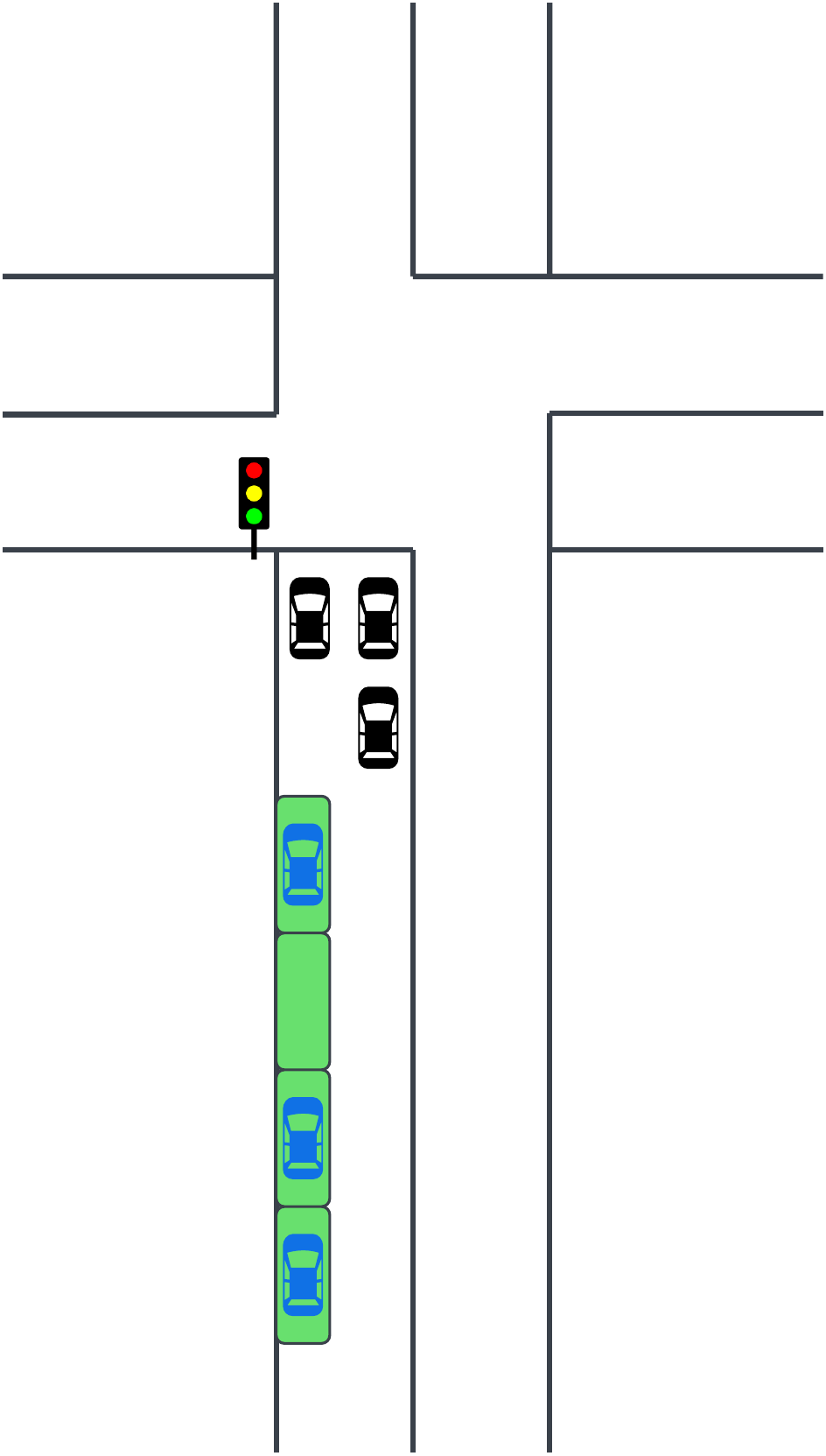}
        \caption{}
        \label{fig:intro_fig4}
    \end{subfigure}
    
    \caption{The impact of on-street parking on traffic flow through an intersection illustrated for left hand traffic. (a) Vehicle queues are imbalanced due to parked vehicles which limits traffic throughput through the intersection. (b) Vehicle queues are balanced when few parking spaces are cleared. (c) Parked vehicles obstructing the traffic flow even at the green phase. (d) Minimal impact from parked vehicles when congestion is less.}
    \label{fig:intro_figure}
\end{figure}

The impact of vehicles parked on the side of a lane on traffic intensifies near intersections, as shown in the example in Figure \ref{fig:intro_figure}. In Figure \ref{fig:intro_fig1}, all parking spaces are occupied, creating an imbalance of vehicle queues, with 2 and 7 vehicles, at the red phase of the traffic light. This imbalance causes reduced traffic throughput through the intersection at the green phase. In Figure \ref{fig:intro_fig2}, two parking spaces have been cleared, resulting in a more balanced vehicle distribution, with 4 and 5 vehicles in each lane. This leads to a higher traffic throughput through the intersection at the green phase. Furthermore, parked vehicles can cause disruptions to the traffic flow, even in the green phase. In Figure \ref{fig:intro_fig3}, the front 3 vehicles, which are trying to turn right, are being blocked by oncoming traffic. The 4th vehicle marked in red wants to go straight, and could have done so, if not for the parked vehicle on the left lane. Hence, all vehicles starting from the 4th vehicle will have to wait a significant amount of time, until the front traffic is cleared. It should be also noted that, when the traffic congestion is low, there is no significant impact from clearing parking spaces, since a minimal number of vehicles would queue at the traffic lights. This is shown in Figure \ref{fig:intro_fig4}. 

\begin{figure}[t]
\centering
\begin{tikzpicture}
\pgfplotsset{width=5.7cm,height=3.7cm,compat=1.3}
\pgfplotsset{
    scale only axis,
    xtick={ 0,1,2,3,4,5},
    xticklabel style={font=\footnotesize},
    yticklabel style={font=\footnotesize},
    xmin=0, xmax=5,
    xlabel={\small No. of cleared parking spaces},
    grid=both,
    legend cell align=left,
    legend style={
    at={(0.6,0.98)},  
        anchor=north,
        font=\footnotesize 
    }
}

\begin{axis}[
  axis y line*=left,
  ymin=40, ymax=120,
  ylabel=\small Time loss (s),
]
\addplot[
    color=blue,
    mark=*,
    thick
] coordinates {
    (0,111.65)
    (1,91.02)
    (2,84.17)
    (3,81.58)
    (4,79.89)
    (5,78.71)
}; \label{plot_one_intro}

\end{axis}

\begin{axis}[
  axis y line*=right,
  axis x line=none,
  ymin=0, ymax=40,
  ylabel=\small Time loss reduction \%
]
\addlegendimage{/pgfplots/refstyle=plot_one_intro}\addlegendentry{Time loss}
\addplot[
    color=red,
    mark=square*,
    thick
] coordinates {
    (0,0.00)
    (1,18.47)
    (2,24.61)
    (3,26.93)
    (4,28.44)
    (5,29.50)
};

\addlegendentry{Time loss reduction \%}
\end{axis}

\end{tikzpicture}
\caption{Subsequent reductions in average travel time loss obtained by clearing parking spaces one by one.}
\label{fi:intro_chart}
\end{figure}
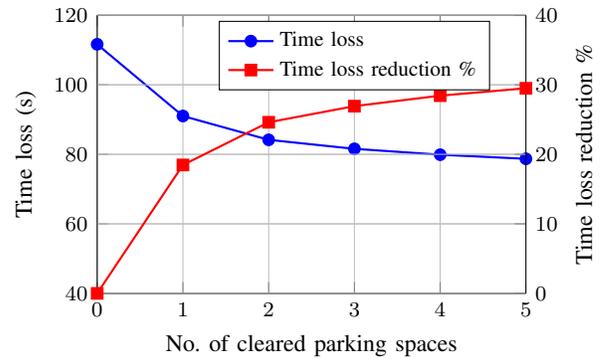

As a solution to the traffic caused by parked vehicles, clearways have been implemented, which completely prevents on-street parking in certain road segments, usually in the rush hours \cite{zhang2011micro}. Clearways is a static solution with well established times and locations, and it does not consider the current traffic conditions or the parking needs of people. We demonstrate why dynamic parking configuration will be highly beneficial over existing clearways using an initial experiment conducted using SUMO \cite{SUMO2018}. Figure \ref{fi:intro_chart} plots the average travel time loss and the average travel time loss reduction percentage (compared with no parking spaces cleared), along the number of cleared on-street parking spaces, on a road segment near an intersection. It could be observed that, while the average travel time loss could be reduced significantly by clearing the first few parking spaces, the subsequent reductions obtained would be essentially incremental. On the other hand, catering to the parking needs is also equally important, because otherwise, vehicles would end up cruising around the neighbourhood looking for a parking space, building up more traffic.

Motivated by the aforementioned observations, we mathematically model the dynamic configuration of on-street parking spaces as an optimization problem with the combined goals of minimizing the average travel time and the walking distance for parking. Dynamic parking space configuration problem is considerably different from other traffic optimization problems such as traffic signal control, since once a car is allowed to park, that action cannot be revised for a certain period of time. Hence, the optimization strategy should be highly proactive and take optimal actions that align with future traffic conditions. Moreover, since the environment changes dynamically, the solution to the optimization problem has to be recomputed with a reasonable frequency. Hence, we follow a data driven approach for obtaining a heuristic solution to the optimization problem, considering the computational complexity associated with other alternative techniques such as linear programming.

To the best of our knowledge, we are the first to look into the problem of dynamically configuring on-street parking spaces and we propose a two-layer multi agent reinforcement learning based solution, considering the nature of the problem. The lower level RL agents are deployed in each lane, which decide the number of parking spaces to be cleared, based on the local observations. Deep Q-learning networks (DQN) \cite{DQN} have been widely used in recent RL based applications, due to their proficiency of approximating the optimal policy in complex environments. However, vanilla DQN does not capture spatial and temporal correlations well, since the observations are given as a single dimensional state vector. We propose an enhanced DQN architecture which captures temporal correlations, i.e., past traffic variation in the respective lane using long short term memory (LSTM) \cite{LSTM} networks, and spatio-temporal correlations from nearby lane segments using graph attention networks (GATs) \cite{GAT}. The resultant state vector has a thorough representation of the environmental conditions, which can be used by the agent to anticipate future traffic conditions. The block level agents (higher level agents) observe the overall parking supply-demand balance around the block, and control the proposed actions by the lane level agents. The goal of the block level agents is to ensure that the parking needs of people are not compromised by the actions of the lane level agents. The local nature of the traffic optimization objective and the neighbourhood nature of the parking behaviour validate the rationale behind our two-layered architecture.

We conduct a series of experiments using the microscopic traffic simulator SUMO \cite{SUMO2018}, in order to justify the value of our proposition. We use both synthetic data as well as real-world data obtained from the records published by the Victorian government for the city of Melbourne. We compare our approach with a set of baselines including clearways, which is the current most widely implemented solution. The results obtained on real-world data provide strong evidence that the average travel time loss of vehicles could be considerably reduced, reaching upto 47\%, at a slight increase (less than two meters) in average walking distance for parking. Vehicle emissions are reduced at the same time, making a positive impact on the environment. Furthermore, we carry out a sensitivity analysis to showcase the robustness and scalability of the proposed solution, and an ablation study, to assess the contribution of each introduced concept. In summary, our contributions are as follows:
\begin{enumerate}
\setlength\itemsep{0.7ex}
    \item We formulate the dynamic configuration of on-street parking spaces, as an optimization problem.
    \item We propose a scalable, two-layered, multi agent reinforcement learning based framework for configuring on-street parking spaces dynamically, based on the observed conditions. Our solution comprises a set of lane level agents acting locally, and a set of block level agents overlooking at the lane level agents.
    \item We introduce a novel Deep Q-learning based reinforcement learning architecture to effectively capture the spatial and temporal variations to learn an optimized policy for lane level agents, that proactively takes decisions which align with future traffic conditions.
    \item We conduct a series of experiments using SUMO, which comprehensively validates the significance of our approach, by achieving a considerable reduction in average travel time loss of vehicles.
\end{enumerate}

\section{Related Work}

Although many earlier works emphasize the impact on traffic congestion caused by on-street parking \cite{guohong,sugiarto,caojin}, optimizing the configuration of parking spaces to minimize the effect on traffic throughput is not a well established area. In recent transport research, a genetic algorithm based optimization is carried out using crossover and mutation procedures in \cite{genetic}, in order to find the optimal configuration of on-street parking. However, their focus is on assigning the whole lane segment as a driving lane or a parking lane, which is less granular than our approach, and can be considered as an extension to the static clearways. Moreover, the genetic optimization algorithm takes a considerable time to converge, even for the considered $5 \times 5$ grid network. A customized advantage actor-critic (A2C) \cite{A2C} algorithm has been used for curbside parking space management in \cite{yincurbside}. However, their intention is to allocate the total number of parking spaces between different types of vehicles rather than a dynamic configuration. The dynamic zoning of parking spaces has been modelled as a mixed integer problem in \cite{nazirzoning}. Solving this optimization problem requires high computation time, and their objective is to zone each parking space as a paid parking space, a loading zone, or a bus stop, which is notably different from our formulation. The following two subsections include recent literature on techniques used for on-street parking optimization and multi agent reinforcement learning based approaches for similar problems in the intelligent transportation systems domain.

\subsection{On-Street Parking Optimization}
\label{literature:parking}

On-street parking has become a scarce resource in most large cities, and it is common for drivers to spend a considerable time in a neighbourhood, looking for a vacant parking spot \cite{SHOUP2006479}. Most of the existing literature has focused on this direction, i.e., assigning a parking space to each driver minimizing driver's cruising time and walking cost, rather than utilizing a parking space dynamically depending on the current traffic conditions.

A multi hop wireless parking meter network (PMNET) is introduced in \cite{basu}, which provides a scalable and efficient way of quickly finding an available parking space closer to the vehicle's current location. The online parking assignment problem is formulated as a mixed-integer linear programming (MILP) problem in \cite{milp}, with the objectives of minimizing cruising time and parking cost. An agile urban parking recommendation service is proposed in \cite{taiwan}, which provides parking lot recommendation sequences, along with successful parking probability values for drivers, in real-time. A receding horizon optimization framework is used in \cite{secon} for recommending on-street parking spots in real-time with the objective of minimizing the driving and walking costs. 

A learn-to-rank based on-street parking recommendation system OPR-LTR, has been introduced in \cite{sun2024},  for recommending a list of ranked parking spaces for a user query. The authors have tested and justified their approach using on-street parking meter data in San Francisco and Hong Kong. The parking search process in a grid is modeled as a Markov Decision Process in \cite{markov}, and a modified Q-learning algorithm has been proposed to achieve faster convergence. All of these approaches are focused on assigning or recommending a parking space to a driver, and their objective is not to reduce travel time globally across all vehicles.

\subsection{Multi Agent Reinforcement Learning in Intelligent Transportation Systems}

Multi agent reinforcement learning (MARL) is a subfield of reinforcement learning, which focuses on developing systems with multiple agents who interact with the environment and take actions individually and collaboratively towards a common goal. Multi agent reinforcement learning has been used to solve a wide variety of applications, and we are specifically interested in the applications in the transport domain. 

Traffic signal control can be given as a key research area where MARL based solutions have been incorporated. Each traffic signal junction is modeled as an individual agent and cost feedback signals from neighbouring agents are used to update the Q-factors in \cite{prabu}, under a multi agent Q-learning strategy. The proposed method has been evaluated on two different traffic settings using the VISSIM \cite{vissim2} traffic simulator. The advantage actor-critic (A2C) algorithm \cite{A2C} has been improved for a multi agent setting in \cite{chumarl} for controlling a multi intersection traffic signal system. Using a set of experiments conducted in the SUMO \cite{SUMO2018} simulation environment, authors justify how the proposed algorithm works better than the existing methodologies. Other notable works which use multi agent reinforcement learning for traffic signal control include \cite{wumarl}, \cite{WANGmarl} and \cite{mamarl}.

A two-layer multi agent reinforcement learning architecture has been incorporated in \cite{gunarathna2021real} for changing lane configuration of roads dynamically based on the traffic patterns. The online parking assignment problem in an indoor parking lot is modeled as a multi agent reinforcement learning problem in \cite{marlparking} and the authors have modified the Monotonic Value Function Factorisation (QMIX) \cite{qmix} algorithm by exploiting the nature of the problem for providing parking bay assignments in a mixed environment of both connected and non-connected vehicles. The aforementioned multi agent reinforcement learning architectures cannot be directly applied to our problem, since the actions have high dependency on spatio-temporal observations, and they should be highly proactive as well. 

\section{Problem Statement}
\label{chap:problem}

The problem of dynamically configuring parking spaces can be formalized as an optimization problem as follows.

\begin{figure}[t]
    \begin{center}
        \includegraphics[width=0.98\linewidth]{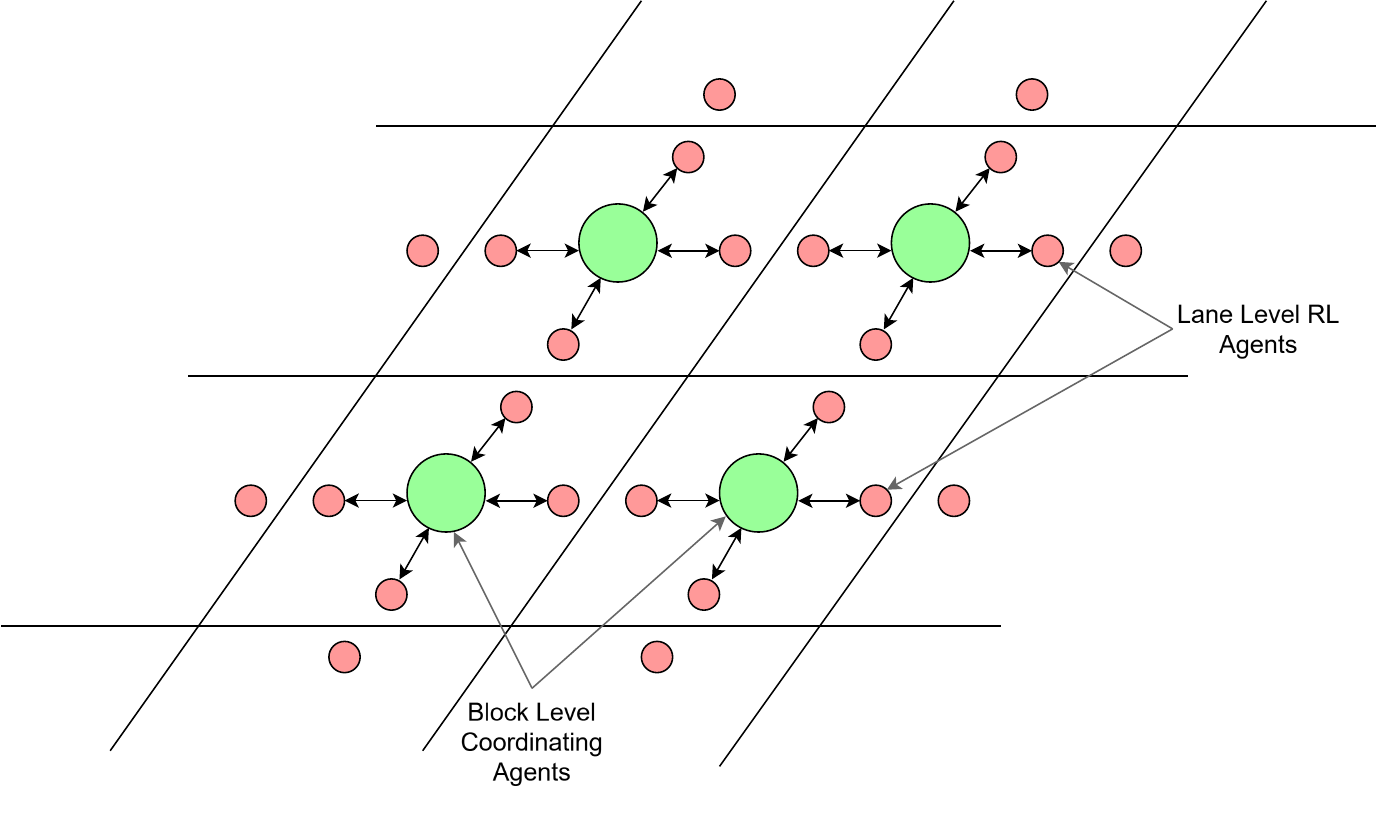}
         \caption{Overview of the proposed architecture. Lane level RL agents determine the parking configuration for each lane, and block level agents overlook them to maintain sufficient parking availability in the neighbourhood.}
        \label{fi:architecture}
    \end{center}
\end{figure}

Given a road network graph $G$, with $K$ number of parking lanes and $M_{j}$ number of parking spaces in lane $j$, let,

\vspace{1em}

\begin{itemize}[itemsep=0.5ex]
    \item[] $N(t)$ - Number of traversing vehicles through $G$ at time $t$
    \item[] $TC_{i}(t)$ - Travel time of vehicle $i$ on $G$ at time $t$
    \item[] $TTC(t)$ - Total travel cost, i.e., $ \sum_{i=1}^{N(t)} TC_{i}(t)$
    \item[] $WC_{i}(t)$ - Walking distance of vehicle $i$ at time $t$
    \item[] $TWC(t)$ - Total walking cost, i.e., $ \sum_{i=1}^{N(t)} WC_{i}(t)$
    \item[] $C_{j}(t)$ - Number of parking spaces cleared in parking lane $j$ at time $t$
    \vspace{1em}
    \item[] Goal is to optimize $C_{j}(t)$ at time $t$ $s.t.$,
    \item[] \hspace{1cm} $\alpha TTC(t) + \beta TWC(t)$ is minimized, where $\alpha$ and $\beta$ are the weighting coefficients.
\end{itemize}

The walking cost, $WC_{i}(t)$ is defined as the distance between the target location and parking location of vehicle $i$. The number of parking spaces cleared in parking lane $j$ at time $t$, $C_{j}(t)$ can vary between $0$ and $M_{j}$, and parking spaces are cleared starting from the ending intersection of lane $j$, since the impact of parked vehicles on traffic intensifies near the intersections, as explained in Section \ref{introduction}. The walking cost is defined as an optimization goal, rather than a constraint, which allows the solution to arrive at an optimal trade-off between the average travel time and the walking distance. It should be also noted that minimizing cruising time has not been set as a direct optimization goal, since the total travel cost, $TTC(t)$ accounts for that indirectly.

 \begin{figure*}[t]
    \begin{center}
        \includegraphics[width=\linewidth]{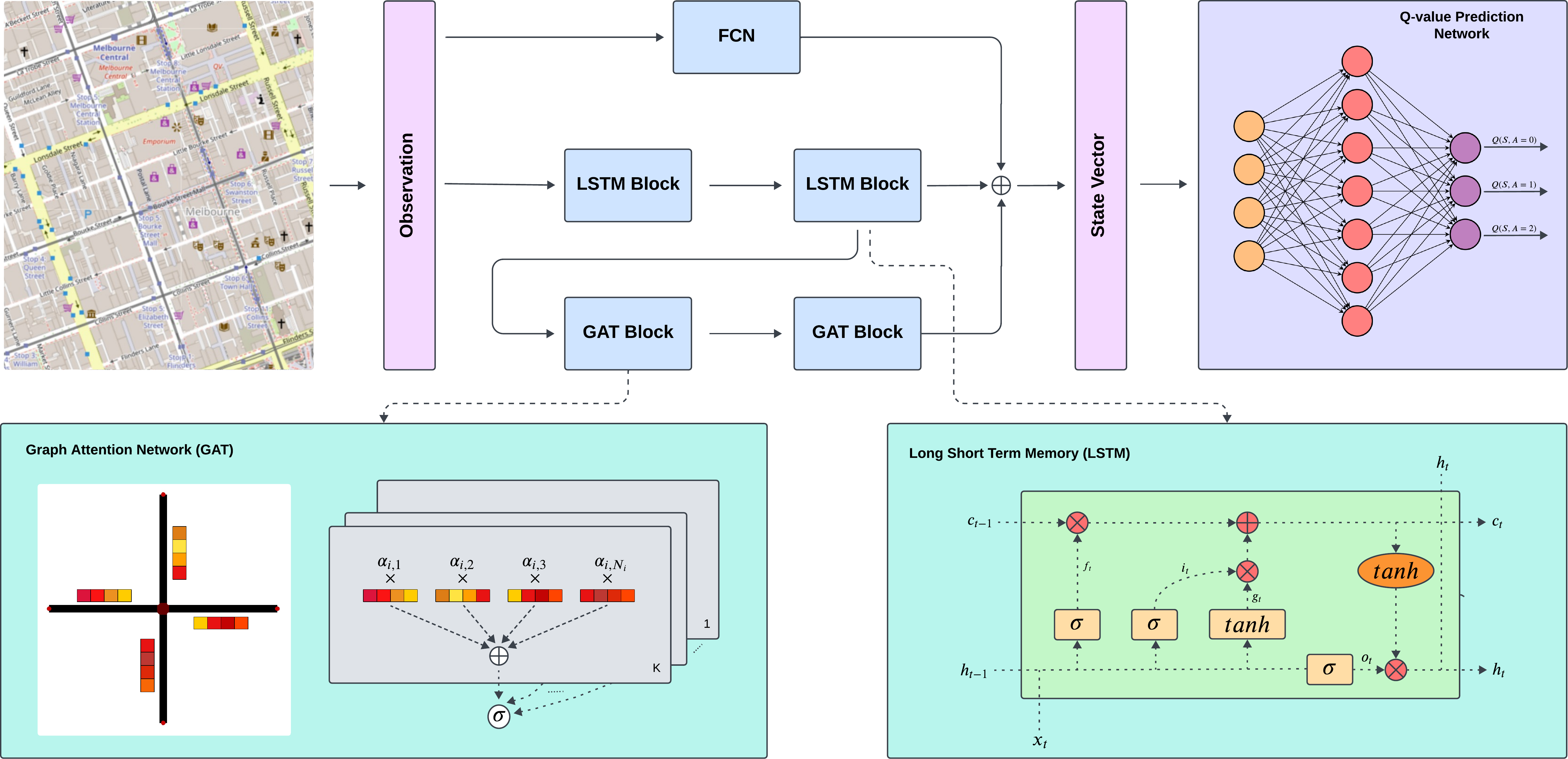}
         \caption{Deep Q-learning network architecture of the lane level agents. The state vector for the Q-value prediction network consists of three main components; linear observational data captured by a fully connected network, the temporal variation of traffic congestion in the lane captured by LSTM blocks, and the variation of traffic congestion of adjacent lane segments captured by LSTM + GAT blocks. The agent selects the action with the highest Q-value for the current state, predicted by the Q-value prediction network.} 
        \label{fi:lane_agent}
    \end{center}
\end{figure*}

\section{Methodology}

To solve the above optimization problem, we propose a novel, dynamic, multi agent reinforcement learning based solution which is explained in detail, in the following subsections.

\subsection{Overall Architecture}

The overall architecture of our proposition is shown in Figure \ref{fi:architecture}. The number of parking spaces to be restricted or allowed in each lane segment is controlled by the lower level RL agents. The top level coordinating agents are responsible for controlling the actions of the lower level agents and managing the parking space availability around a block. The resultant hierarchical framework is inherently scalable for large road networks with multiple lane segments.

\subsection{Lane Level RL Agents}

The optimal parking space configuration for a lane is decided by lane level agents, based on the observed environmental conditions.

\subsubsection{RL Problem Formulation}

Each lane level agent is modelled as a RL agent that executes one of the three actions; increasing the number of cleared parking spaces by 1, reducing the number of cleared parking spaces by 1 or keeping the current configuration. The optimal action should be taken considering the future traffic conditions and a holistic picture of the local environment is obtained by including the current parking space configuration, the current level of traffic congestion in the lane, the variation of the traffic congestion of the lane at past time steps, and the variation of the traffic congestion of adjacent lane segments at past time steps as states. The traffic congestion is represented by dividing the number of traversing vehicles on the lane divided by the lane length. By setting a positive reward for reducing the average travel time (to minimize $TTC(t)$) and a negative reward for increasing the walking distance (to minimize $TWC(t)$), the agent is trained to find a balance between the two extremes of removing all parking spaces and keeping all parking spaces for the observed environmental conditions. The coefficients $\alpha$ and $\beta$ are used as weights in the reward function for the two components, and they are tuned as hyper-parameters while training the RL model.

\subsubsection{Deep Q-learning Network Architecture}

For training the lane level agents, we introduce a novel Deep Q-learning framework to effectively capture the spatio-temporal correlations present in our problem, as shown in Figure \ref{fi:lane_agent}. 

\subsubsection{Fully Connected Network}

 We embed the linear observational data into a latent space vector using a fully connected layer, as follows.

\begin{equation}
   v_{fcn, i}^t = ReLU(o_{c,i}^tW_c + b_c)
\end{equation}

Here, \( o_{c,i}^t \) denotes the observation of lane $i$ at time $t$, where \( W_c\) and \( b_c\) are the learnable weight matrix and bias vector of the fully connected network. The output embedding vector $v_{fcn}^t$ represents the current state of the lane $i$ at time $t$.

\subsubsection{Long Short Term Memory Networks}

The second embedding vector is created by applying a LSTM \cite{LSTM} network on the observed traffic congestion values over a period of time to capture the temporal behaviour of traffic. We use a two-layered LSTM network and the input for the second layer is obtained by multiplying the hidden states of the first layer by a dropout layer. The whole operation of the LSTM network can be summarized as follows, where $o_{l,i}^t$ denotes the observation vector of traffic congestion of lane $i$ over a period of time and $v_{lstm,i}^t$ denotes the output embedding vector of $i^{th}$ lane at time $t$.

\vspace{-0.3em}
\begin{equation}
   v_{lstm,i}^t = LSTM(o_{l,i}^t)
\end{equation}
\vspace{0.3em}

\subsubsection{Graph Attention Networks}

Furthermore, a graph attention network \cite{GAT} based architecture is used to obtain an understanding of the traffic conditions of the adjacent lanes. The observed traffic congestion values of the adjacent lanes in a period of time are first passed through the same LSTM network to get feature representations for each lane $k$, $x_{k}$. Then the importance of features of lane $j$ on lane $i$ is then calculated as per the following operation. $W_{s}$ and $W_{t}$ denote the embedding parameters, and $a_s$ and $a_t$ denote the learnable attention vectors for source and target lanes, respectively.

\begin{equation}
    e_{ij} = LeakyReLU(a_s^T W_s x_i + a_t^T W_t x_j)
\end{equation}

Then the obtained interaction scores are normalized within a selected neighbourhood $N_{i}$ of the lane $i$, using the softmax function as follows to obtain the attention coefficient $\alpha_{ij}$.

\begin{equation}
\alpha_{ij} = softmax(e_{ij}) = \frac{\exp(e_{ij})}{\sum_{k \in \mathcal{N}(i)} \exp (e_{ik})} 
\end{equation}

\begin{algorithm}[!t]
\small
\DontPrintSemicolon
  
  \KwInput{$t_{l}$ - Time between two lane level agent actions}
    \KwInput{$t_{b}$ - Time between two block level agent actions}
  \KwInput{$th_{o}$ - Parking space occupancy threshold}
  \KwInput{$th_{c}$ - Number of cruising vehicles threshold}
    $t_{step} \leftarrow 0$ \\
    \While{True}
    {
        \If{$t_{step} \hspace{1mm} mod \hspace{1mm} 
 t_{l} = 0$}
        {
            \ForEach{$agent \in lane \hspace{1mm} agents$}    
            { 
                Determine the optimal action for the respective lane based on the observed conditions \\
            }
            \ForEach{$agent \in block \hspace{1mm} agents$}    
            { 
                $O_{block} \leftarrow $ Parking occupancy of the block \\
                \ForEach{$lane \in respective \hspace{1mm} block$}
                {
                \If{$suggested \hspace{1mm} action = increase$}
                {\If{$O_{block} < th_{o}$}{
                    Execute the action \\
                }
                    
                }
                \Else{
                    Execute the action \\
                }
            }
            }
            
        }
        \If{$t_{step} \hspace{1mm} mod \hspace{1mm} t_{b} = 0$}
        {
                    $V_{c} \leftarrow $ Number of cruising vehicles in the block \\
            \If{$V_{c} > th_{c}$}{
            \ForEach{$lane \in respective \hspace{1mm} block$}{
                Allow more parking spaces to be utilized on the respective lane based on $V_{c}$ \\ 
            }
            }
        }
    $t_{step} \leftarrow t_{step} + 1$
    }

\caption{Dynamic on-street parking configuration algorithm. Each lane level agent observes the environmental conditions, and determines the best action for the respective lane. The block level agents will allow or restrict the action based on the observed parking occupancy. In addition, block level agents monitor the number of cruising vehicles, and allow more parking spaces to be utilized, accordingly.}
\label{algo_1}
\end{algorithm}

The obtained normalized attention coefficients are then combined together to model the overall influence of neighbouring lanes on the target lane.

\begin{equation}
h_{i} = ReLU ( \sum_{j \in \mathcal{N}_i} \alpha_{ij} W_t x_j)
\end{equation}

The above operations are performed in parallel with different sets of learnable parameters ($W_t$, $W_s$, $a_t$, $a_s$) as multiple attention heads, and the output is concatenated as follows.

\begin{equation}
    hm_i = ReLU( \big\|_{m=1}^M  \sum_{j \in \mathcal{N}_i} \alpha_{ij}^m W_t^m x_j )
\end{equation}

The output is then passed through another multi-head graph attention network layer to capture multi hop information and learn richer node representations. At the output layer, we obtain the average over multiple attention heads instead of the concatenation operation as follows.

\begin{equation}
hm_{i} = ReLU(\frac{1}{M} \sum_{m=1}^M \sum_{j \in N_{i}} \alpha_{ij}^{m} W_{t}^m x_j)
\end{equation}

In summary, the embedding vector from the graph attention network, $v_{gat,i}^t$ for lane $i$ at time $t$, is obtained as follows.

\begin{equation}
   x_{i}^t, x_j^t, ... , x_K^t = LSTM(o_{l,i}^t, o_{l,j}^t, ... , o_{l,K}^t)
\end{equation}
\begin{equation}
   hm_{i}^t, hm_j^t, ... , hm_K^t = GAT(x_{i}^t, x_j^t, ... , x_K^t)
\end{equation}
\begin{equation}
   v_{gat,i}^t = GAT(hm_{i}^t, hm_j^t, ... , hm_K^t)
\end{equation}

\begin{algorithm}[t]
\small
\DontPrintSemicolon
  
  \KwInput{$P_{t}$ - Target parking space}
  \KwInput{$th_{d}$ - Walking distance threshold}
  \KwInput{$A_{p}$ - Array of parking spaces}
  \KwInput{$G$ - Road network}
  $parked \leftarrow $ False \\
  \While{$parked = False$}{
    \If{$P_{t} = Available$}{
        Proceed to park in $P_{t}$ \\
        $parked \leftarrow $ True \\
    }
    \Else{
    $A_{ps} \leftarrow $ Sort array of parking spaces ($A_{p})$ based on the distance from the target parking space (from smallest to the largest)\\
    }
    \ForEach{$parking \hspace{1mm} space \hspace{1mm} P_{i} \in A_{ps}$}{
        \If{$P_{i} = Available$}{
            Proceed to park in $P_{i}$ \\
            $parked \leftarrow $ True \\
            $break$ \\
        }
        \If {$dist(P_{i}, P_{t}) > th_{d}$}{
            $break$ \\
        }
    }
    \If{$parked = False$}{
        Cruise around the block
    }
    }
\caption{The parking search algorithm. Each vehicle checks whether its target parking space is available, and if not, it searches for a vacant parking space within the walking distance threshold, starting from the closest. If no parking space is found within a close vicinity, it would cruise around.}
\label{algo_2}
\end{algorithm}

\subsubsection{Q-value Prediction Network}

A three layer fully connected network is used as the Q-value prediction network, and the input state vector is obtained by concatenating the three embedding vectors, as follows. 

\begin{equation}
   v_{state,i}^t = v_{fcn,i}^t \| v_{lstm,i}^t \| v_{gat,i}^t 
\end{equation}

The cost function minimized at training is described using Equations \ref{eq:dqn1}, \ref{eq:dqn2} and \ref{eq:dqn3}, where $\tilde{\theta}$ and $\tilde{\theta}$ denote the parameters of the online and target networks used with DQN.

\begin{equation}
L = \frac{1}{N} \sum_{n=1}^{N} \left( y_n - Q(S_{n}, A_{n}, \theta) \right)^2
\label{eq:dqn1}
\end{equation}

\noindent where if the next state \( S_{n+1} \) is not the terminal state:
\begin{equation}
y_n = R_{n} + \gamma \max_A Q(S_{n+1}, A, \tilde{\theta})
\label{eq:dqn2}
\end{equation}

\noindent and if the next state \( S_{n+1} \) is the terminal state:
\begin{equation}
y_n = R_{n}
\label{eq:dqn3}
\end{equation}

The Q-value prediction network predicts Q-values for each action, and the agent selects the action with the highest Q-value. 

\subsubsection{Multi Agent Reinforcement Learning Setup}

Due to the homogeneous nature of our problem, each lane agent has a similar local optimization problem. Hence we deploy homogeneous RL agents with a common trained policy in each lane, following the centralized training with decentralized execution approach in multi agent reinforcement learning.

\subsection{Block Level Agents}

The block level agents are responsible for managing the actions of the lower level agents, with the goal of providing sufficient parking availability around a block. The block level agents achieve this in two ways. First, when a lower level lane agent suggests to increase the number of cleared parking spaces by one, it will analyse the current level of parking occupancy in the block, and will only allow the change to proceed if the occupancy is less than a given threshold. Second, it continuously analyses the number of cruising vehicles around the block, and if that exceeds a given threshold, it will release parking spaces cleared by the lane level agents, accordingly. Since vehicles typically search for parking within a confined geographic area, utilizing higher level agents at the block level is sufficient, to ensure that the actions of the lane level agents will not compromise the parking needs of people.

\subsection{Algorithms}

The dynamic parking space configuration process is explained in Algorithm \ref{algo_1}. Each lane level agent determines the best action for the respective lane using the trained RL model, based on the observed environmental conditions. The proposed action could be either to increase the number of cleared parking spaces by 1, or to reduce the number of cleared parking spaces by 1, or to keep the current configuration. The block level agents observe the parking space occupancy of the block, and execute the each lane level action accordingly. If the proposed lane level action is to increase the number of cleared parking spaces, the action would only get executed if the parking space occupancy of the respective block is less than a given threshold. In addition to that, lane level agents monitor the number of cruising vehicles around the block at a higher frequency, and if that number gets higher than a given threshold, it will allow more parking spaces to be utilized in each lane of the block, based on the number of cruising vehicles. The parameter setup for the dynamic parking space configuration algorithm is presented in Table \ref{ta:common_parameters}. 

Algorithm \ref{algo_2} explains how parking spaces are allocated to vehicles. Each vehicle has a target parking space, which is the closest to its destination. However, the parking space might not be available due to being restricted by a policy or being occupied by another vehicle. If so, it would then consider other nearby parking spaces one by one, starting from the closest, within the acceptable walking distance threshold. In the event of no available parking space within the considered region, the vehicle would cruise around the block until a parking space becomes available.

\begin{figure}[t]
    \begin{center}
        \includegraphics[width=0.87\linewidth]{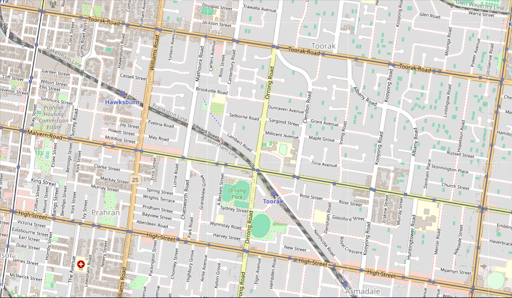}
         \caption{Melbourne suburb region used in the experiments covering 15 intersections, 38 lane segments and 3042 parking spaces.}
        \label{fi:osm}
    \end{center}
\end{figure}

\begin{table}[t]
\centering
\centering
\begin{tblr}{
  width = 0.99\linewidth,
  colspec = {Q[6.25cm]Q[1.2cm]},
  column{2} = {c},
  row{1} = {font=\bfseries},
  hline{1,2,6},
  vline{2},
}
Parameter    & Value \\
Time between two lane level agent actions ($t_{l}$)       & 100 s  \\
Time between two block level agent actions ($t_{b}$)      & 10 s   \\
Parking bay occupancy threshold ($th_{o}$)      & 0.8   \\
Number of cruising vehicles threshold ($th_{c}$)      & 4     \\
\end{tblr}
\vspace{1em}
\caption{Parameter setup for D-PA algorithm}
\label{ta:common_parameters}
\vspace{-1em}
\end{table}

\begin{table}[t]
\centering
\begin{tblr}{
  width = \linewidth,
  colspec = {Q[l,m,3cm] Q[m,m,4.5cm]},
  column{2} = {c},
  row{1} = {font=\bfseries},
  hline{1,2,6},
  vline{2},
}
Parameter & Value Range\\
Vehicle insertion rate (veh/s)  & \{60, 70, 80, \textbf{90}, 100, 110\}\\
Parking probability & \{0.1, 0.15, \textbf{0.2}, 0.25, 0.3, 0.35, 0.4\}\\
Parking duration (s)  & \{\textbf{600}, 1200, 1800, 2400, 3000, 3600\}\\
Grid size & \{\textbf{3} {\boldmath $\times$} \textbf{3}, 4 $\times $ 4, 5 $\times $ 5, 6 $\times$ 6, 7 $\times$ 7\}\\
\end{tblr}
\vspace{1em}
\caption{Parameter setup for sensitivity analysis. The default value is emphasized in bold, within the considered value range for each parameter.}
\label{ta:paramters_sensitivity}
\vspace{-1em}
\end{table}

\section{Experiments}

We conduct a series of experiments to investigate the benefits of the proposed system using the SUMO \cite{SUMO2018} simulation environment. The Python API TraCI available in SUMO, is leveraged for integrating the multiagent system with the simulator. The environmental conditions such as the traffic congestion levels are retrieved via TraCI at the simulation runtime. Then the trained RL agents along with the block level agents determine the best courses of action for each lane, which are directly executed using TraCI. 

\begin{table*}[t]
\centering
\centering
\small
\begin{tblr}{
  colspec = {Q[0.13\textwidth] Q[0.047\textwidth] Q[0.047\textwidth] Q[0.047\textwidth] Q[0.047\textwidth] Q[0.047\textwidth] Q[0.047\textwidth] Q[0.047\textwidth] Q[0.047\textwidth] Q[0.047\textwidth] Q[0.047\textwidth] Q[0.047\textwidth] Q[0.047\textwidth]},
  rows = {c,m},
  column{1}={halign=l},
  cell{1}{1} = {r=3}{c},
  cell{1}{2} = {c=12}{c},
  cell{2}{2} = {c=3}{c},
  cell{2}{5} = {c=3}{c},
  cell{2}{8} = {c=3}{c},
  cell{2}{11} = {c=3}{c},
  hlines,
  vline{2,5,8,11} = {-}{},
}
Baseline & Parking Probability &       &       &       &       &       &       &  & & & &    \\
         & 0.1 & & & 0.2 & & & 0.3 & & & 0.4 & &   \\
& {$t_{loss}\downarrow$ \\ $\text{\footnotesize (s)}$}  & {$t_{loss}\uparrow$ \\ $  \text{\footnotesize (\%)}$} & {\text{\footnotesize $d_{walk}\downarrow$} \\ $\text{\footnotesize (m)}$}
& {$t_{loss}\downarrow$ \\ $\text{\footnotesize (s)}$}  & {$t_{loss}\uparrow$ \\ $ \text{\footnotesize (\%)}$} & {\text{\footnotesize $d_{walk}\downarrow$} \\ $\text{\footnotesize (m)}$}
& {$t_{loss}\downarrow$ \\ $\text{\footnotesize (s)}$}  & {$t_{loss}\uparrow$ \\ $ \text{\footnotesize (\%)}$} & {\text{\footnotesize $d_{walk}\downarrow$} \\ $\text{\footnotesize (m)}$}
& {$t_{loss}\downarrow$ \\ $\text{\footnotesize (s)}$}  & {$t_{loss}\uparrow$ \\ $ \text{\footnotesize (\%)}$} & {\text{\footnotesize $d_{walk}\downarrow$} \\ $\text{\footnotesize (m)}$}\\

No-PA & 113.85 & - & \textbf{0.48} & 145.14 & - & \textbf{0.98} & 190.84 & - & \textbf{1.49} & 232.17 & - & \textbf{2.12} \\
S-PA & 106.84 & 6.16\% & 0.80 & 133.10 & 8.30\% & 1.29 & 169.59 & 11.14\% & 1.71 & 200.21 & 13.77\% & 2.47 \\
C-PA & 109.65 & 3.69\% & 59.75 & 129.82 & 10.56\% & 63.52 & 167.40 & 12.28\% & 68.09 & 198.55 & 14.48\% & 70.70 \\
D-PA (PPO) & 108.80 & 4.44\% & 0.72 & 129.49 & 10.78\% & 1.29 & 150.25 & 21.27\% & 1.86 & 169.90 & 26.82\% & 2.56 \\
D-PA (A2C) & 106.92 & 6.09\% & 0.89 & 125.34 & 13.64\% & 1.50 & 144.22 & 24.43\% & 2.13 & 167.22 & 27.98\% & 2.86 \\
D-PA (DQN) & 105.36 & 7.46\% & 0.85 & 121.05 & 16.60\% & 1.43 & 138.41 & 27.47\% & 2.00 & 155.71 & 32.93\% & 2.72 \\
D-PA (D-DQN) & 109.43 & 3.89\% & 0.71 & 131.82 & 9.18\% & 1.27 & 155.19 & 18.68\% & 1.83 & 176.65 & 23.91\% & 2.54 \\
D-PA (Du-DQN) & 108.69 & 4.54\% & 0.71 & 129.07 & 11.07\% & 1.28 & 153.66 & 19.48\% & 1.85 & 170.06 & 26.75\% & 2.58 \\
D-PA (Ours) & \textbf{99.02} & \textbf{13.03\%} & 1.44 & \textbf{108.02} & \textbf{25.58\%} & 2.11 & \textbf{115.07} & \textbf{39.70\%} & 2.71 & \textbf{121.22} & \textbf{47.79\%} & 3.52
\end{tblr}
\vspace{1em}
\caption{Experimental results on real-world data. The proposed D-PA architecture has significantly improved traffic flow with each parking probability, by achieving the lowest average travel time loss and the highest travel time loss reduction percentage, when compared with all other baselines. The walking distance has only increased by less than two meters in all cases.}
\label{ta:real_world}
\end{table*}

\subsection{Data}

Both synthetic data and real-world data from the city of Melbourne are used for our experiments, as described below.

\subsubsection{Real-World Data}

For the main set of experiments, an OpenStreetMap (OSM) map of a central suburb region in Melbourne, as shown in Figure \ref{fi:osm}, is used to generate the network file for SUMO. This region includes 15 signalized intersections, 38 road segments and 3042 on-street parking spaces, covering a $3.5 km \times 2 km$ area, approximately. Then the inductor loop counter (SCATS) data \cite{scats} and the traffic signal configuration data sheets \cite{datasheets} published by the Victorian Government are joined to obtain the traffic signal volumes at each intersection of the considered region. The routesampler tool in SUMO is then used to generate vehicle trips to match the inductor loop counts which would approximate the actual trips on a given day. The distribution of parking duration is analysed using parking space sensor data \cite{parking_data}, and an exponential distribution is approximated for the parking duration, which is used to extract individual parking durations for each vehicle. A total of 71,217 vehicles representing daily traffic from 6 a.m. to 9 p.m. have been simulated with different parking probabilities.

\subsubsection{Synthetic Data}

In addition to the experiments conducted using real-world data, a sensitivity analysis is carried out using synthetic data, in order to understand the performance of the proposed framework under different conditions. For this set of experiments, a grid network is used, and vehicle trips are created from sources to destinations following a random traffic behaviour. The vehicle insertion rate, parking probability, parking duration and grid size are changed as parameters, to demonstrate the robustness and the scalability of our approach. The parameter setup for the sensitivity analysis is given in Table \ref{ta:paramters_sensitivity}, including the value range and the default value in bold, for each parameter. 

\subsection{Model Training}

PyTorch \cite{pytorch} deep learning framework is used to train the DQN model end-to-end using the Adam optimizer and the mean squared loss (MSE) defined under Equation \ref{eq:dqn1}. The fully connected network used to embed current observational data \( o_{c,i}^t \) into $v_{fcn,i}^t$ has 2 input nodes and 32 output nodes. The LSTM network has 2 layers with a 32 dimensional hidden state, and a sequence length of 10 is used to denote the traffic variation over time \( o_{c,l}^t \). The graph attention networks consist of 2 attention heads and a 64 dimensional hidden state. All three embedding vectors $v_{fcn,i}^t$,  $ v_{lstm,i}^t $ and $v_{gat,i}^t $ are 32 dimensional to give equal importance to the three components of the state vector. The input layer of the Q-value prediction network consists of 96 nodes which takes $v_{state,i}^t$ as the input. The final layer consists of 3 nodes to represent the three actions while the hidden layer consists of 128 nodes. The model parameters are carefully selected and refined for optimal performance.

\subsection{Baselines}

To the best of our knowledge, dynamic configuration of on-street parking spaces is first investigated by us and hence, we compare our approach with the following baselines.
\begin{itemize}
   \item \textbf{No Parking space Allocations (No-PA)} - No parking space restrictions are used.
   \item \textbf{Clearways based Parking space Allocations (C-PA)} - Clearways are used in rush hours, and the times and the locations of the clearways for the real-world experiments are obtained from \cite{clearway_data}. With synthetic data, clearways are applied in one direction at all times.
   \item \textbf{Static Parking space Allocations (S-PA)} - A fixed amount of parking spaces near intersections are cleared at all times in a static manner.
    \item \textbf{Dynamic Parking space Allocations (D-PA)} - Parking spaces near intersections are cleared dynamically, based on the observed environmental conditions. Based on the methodology for clearing parking spaces, this leads to multiple baselines, and we compare our approach with other well-known reinforcement learning model architectures including PPO \cite{ppo}, A2C \cite{A2C}, Double DQN \cite{ddqn} and Dueling DQN \cite{dudqn}.
 \end{itemize}

 \subsection{Evaluation Metrics}

 We evaluate the performance of our proposition and compare with the baselines using the following evaluation metrics.
 
\begin{figure}[t]
    \centering
    \includegraphics[width = 0.97\linewidth]{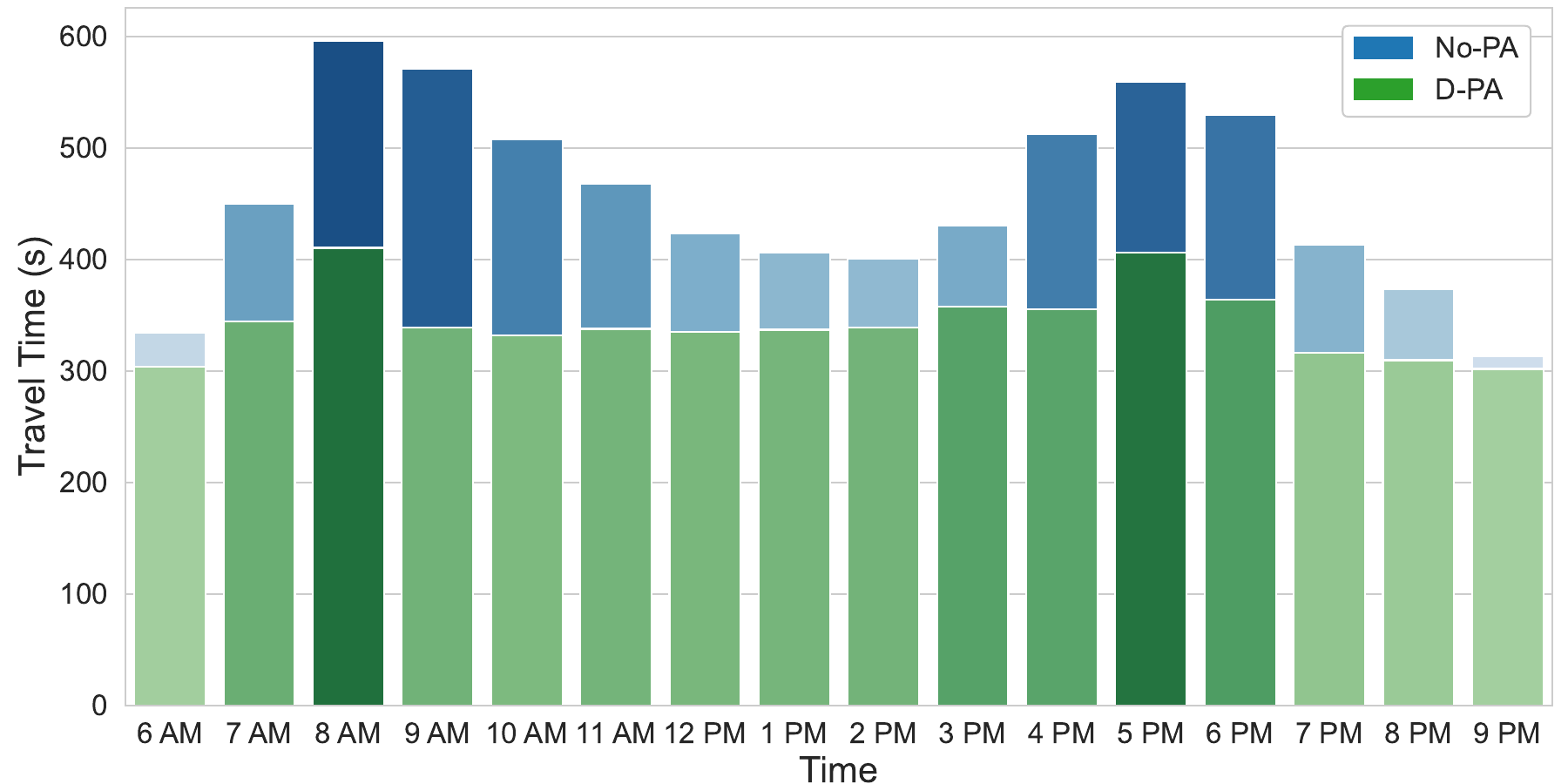} 
    \caption{Hourly distribution of travel times in real-world experiments, with and without D-PA. The traffic flow has been considerably improved at all times by utilizing D-PA, especially during rush hours.}
    \vspace{1em}
    \label{fig:hourly_distribution}
\end{figure}

\begin{itemize}
\setlength\itemsep{1ex}
   \item \textbf{Average Time Loss ($t_{loss}$)} - The time loss is defined as the difference between the actual travel time and the free flow travel time which is the theoretical minimum possible travel time of a vehicle.

    \item \textbf{Average Time Loss Percentage Deviation ($t_{loss} \hspace{1mm} \text{\small \%})$} - This metric is calculated as the difference between the average time loss in the considered baseline with the No-PA baseline as a percentage deviation. 

   \item \textbf{Average Walking Distance ($d_{walk}$)} - Walking distance is the difference between the target destination of a vehicle and the actual parking location.
   
 \end{itemize}

In addition to the above metrics, the gas emissions of the vehicles and the distribution of travel times are studied in detail.

 \section{Results}

 We present our experimental results in this section, including results on real-world data, the sensitivity analysis and the ablation study.

 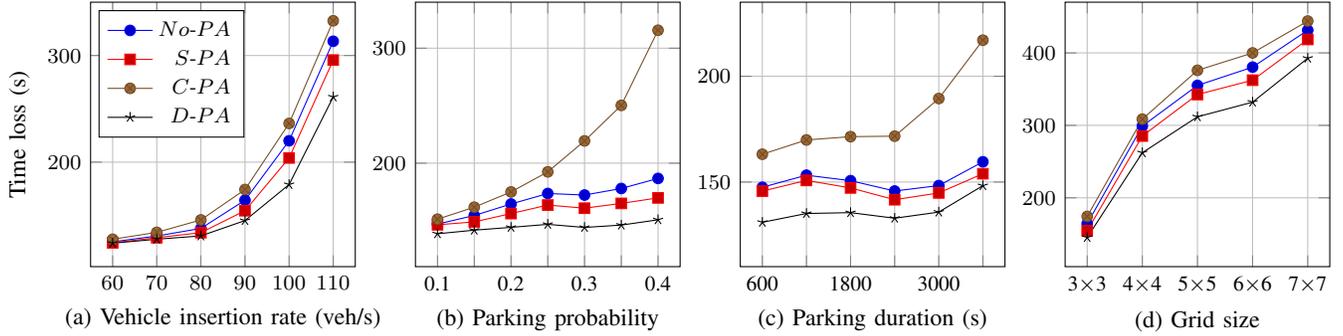
\begin{figure*}
\centering
\begin{tikzpicture}
\begin{groupplot}[
    group style={
        group size=4 by 1,   
        horizontal sep=0.8cm 
    },
    width=5.1cm,   
    height=5.1cm,
    grid=both,
    every axis/.append style={
        label style={font=\small},
        tick label style={font=\footnotesize},
        title style={}
    },
]

\nextgroupplot[xlabel=(a) Vehicle insertion rate (veh/s), ylabel={Time loss (s)}, xtick = data, ytick align=inside, ymin=102,
    ymax=350, legend style={
cells={anchor=east},
legend pos=north west,
font=\footnotesize
}]
\addplot+ coordinates {(60, 125.309) (70, 130.73)  (80, 137.84 )  (90, 164.306) (100, 220.046) (110, 313.343)};
\addplot+ coordinates {(60, 124.349) (70, 129.174)  (80, 133.866)  (90, 154.466) (100, 203.921) (110, 295.725)};
\addplot+ coordinates {(60,	127.671) (70, 134.15)  (80, 145.814)  (90, 174.326) (100, 236.294) (110, 332.565)};
\addplot+ coordinates {(60,	124.005) (70, 127.728)  (80, 130.837)  (90, 145.218) (100, 179.009) (110, 261.164)};
\legend{$No{\text -}PA$,$S{\text -}PA$,$C{\text -}PA$,$D{\text -}PA$}

\nextgroupplot[xlabel=(b) Parking probability, xtick={0.1, 0.15, 0.2, 0.25, 0.3, 0.35, 0.4}, xticklabels={0.1,, 0.2,, 0.3,, 0.4}, xticklabel style={font=\footnotesize}, ytick align=inside, ymin=110,
    ymax=340,] 
\addplot+ coordinates {(0.1, 147.032) (0.15, 154.472)  (0.2, 164.495)  (0.25, 173.623) (0.3, 172.264) (0.35, 177.964) (0.4, 186.732)};
\addplot+ coordinates {(0.1, 146.408) (0.15, 148.933)  (0.2, 156.157)  (0.25, 163.562) (0.3, 160.838) (0.35, 165.009) (0.4, 169.804)};
\addplot+ coordinates {(0.1, 151.257) (0.15, 161.719)  (0.2, 174.984)  (0.25, 192.434) (0.3, 219.366) (0.35, 250.244) (0.4, 315.604)};
\addplot+ coordinates {(0.1, 138.607) (0.15, 141.729)  (0.2, 144.176)  (0.25, 146.783) (0.3, 144.097) (0.35, 146.143) (0.4, 150.756)};

\nextgroupplot[xlabel=(c) Parking duration (s), xtick={600,1200,1800,2400,3000,3600},
  xticklabels={600,,1800,,3000,}, xticklabel style={font=\footnotesize}, ytick align=inside, ymin=110,
    ymax=235,] 
\addplot+ coordinates {(600, 147.515) (1200, 153.223)  (1800, 150.6347)  (2400, 145.7491) (3000, 148.3225) (3600, 159.559)};
\addplot+ coordinates {(600, 145.708) (1200, 150.754)  (1800, 147.212)  (2400, 141.6321) (3000, 144.8476) (3600, 153.9162)};
\addplot+ coordinates {(600, 163.128) (1200, 169.882)  (1800, 171.446)  (2400, 171.667) (3000, 189.462) (3600, 217.0443)};
\addplot+ coordinates {(600, 130.952) (1200, 135.125)  (1800, 135.5259)  (2400, 132.887) (3000, 135.7813) (3600, 148.289)};

\nextgroupplot[xlabel=(d) Grid size, symbolic x coords={3$\times$3,4$\times$4,5$\times$5,6$\times$6,7$\times$7}, xtick = data, ytick align=inside, ymin=105,
    ymax=470,] 
\addplot+ coordinates {(3$\times$3, 164.306) (4$\times$4, 298.99)  (5$\times$5, 355.13)  (6$\times$6, 380.35) (7$\times$7, 431.33)};
\addplot+ coordinates {(3$\times$3, 154.466) (4$\times$4, 285.25)  (5$\times$5, 342.59)  (6$\times$6, 362.34) (7$\times$7, 418.57)};
\addplot+ coordinates {(3$\times$3, 174.326) (4$\times$4, 308.55)  (5$\times$5, 375.95)  (6$\times$6, 399.95) (7$\times$7, 443.94)};
\addplot+ coordinates {(3$\times$3, 145.218) (4$\times$4, 262.21)  (5$\times$5, 311.82)  (6$\times$6, 332.04) (7$\times$7, 392.68)};

\end{groupplot}
\end{tikzpicture}

\caption{Sensitivity analysis results with regard to (a) vehicle insertion rate, (b) parking probability, (c) parking duration, and (d) grid size. The proposed D-PA framework shows robust performance with increasing traffic, parking probability, parking duration and grid size.}
\label{fi:sensitivity_analysis}
\end{figure*}

\begin{figure}[t]
    \begin{center}
        \begin{tikzpicture}
\begin{axis}[
every axis label/.append style={font=\small}, 
every axis tick/.append style={font=\small}, 
ytick scale label code/.code={},
y label style={at={(axis description cs:-0.15,0.5)}, anchor=center},
symbolic x coords={0.1, 0.15, 0.2, 0.25, 0.3, 0.35, 0.4},
height=7.4cm,
width=7.5cm,
grid=both,
legend pos=north west,
legend style={font=\small},
every axis plot/.append style={thick},
xlabel={Parking probability},
ylabel={Reduction percentage},
]

\addplot [color=blue, mark = *] coordinates {
    (0.1, 4.05) (0.15, 6.48) (0.2, 11.21) (0.25, 12.56) (0.3, 15.49) (0.35, 17.81) (0.4, 20.42)
};
\addlegendentry{CO2}

\addplot [color=red, mark = square*] coordinates {
    (0.1, 11.49) (0.15, 17.14) (0.2, 27.13) (0.25, 29.08) (0.3, 34.36) (0.35, 37.90) (0.4, 41.67)
};
\addlegendentry{CO}

\addplot [color=violet, mark = triangle*] 
 coordinates {
    (0.1, 10.49) (0.15, 15.80) (0.2, 25.25) (0.25, 27.20) (0.3, 32.28) (0.35, 35.76) (0.4, 39.48)
};
\addlegendentry{HC}

\addplot [color=brown, mark = diamond*]  coordinates {
    (0.1, 7.42) (0.15, 11.46) (0.2, 18.92) (0.25, 20.65) (0.3, 24.89) (0.35, 27.96) (0.4, 31.35)
};
\addlegendentry{PMx}

\addplot [color=olive, mark = star]  
 coordinates {
    (0.1, 5.02) (0.15, 7.94) (0.2, 13.54) (0.25, 15.04) (0.3, 18.42) (0.35, 21.02) (0.4, 23.92)
};
\addlegendentry{NOx}
\end{axis}
\end{tikzpicture}
         \caption{Analysis on vehicle gas emissions reductions obtained by D-PA including carbon dioxide, carbon monoxide, hydrocarbons, particular materials and nitrogen oxides.}
        \label{fi:emissions}
    \end{center}
\end{figure}
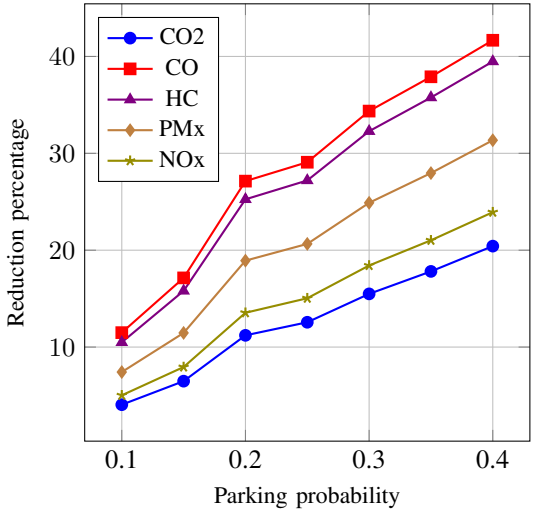

\subsection{Real-World Data}

\subsubsection{Core Evaluation Metrics}

 The experimental results obtained with real-world data are listed in Table \ref{ta:real_world}. Our proposed dynamic parking space configuration architecture has been able to achieve the lowest average travel time loss and the highest travel time loss percentage deviation in all cases, reaching a maximum of 47.79\%, at an increase in walking distance below two meters. In essence, the traffic flow has significantly improved benefiting all traversing vehicles, at a negligible increase in the walking distance for parking. This is further illustrated using Figure \ref{fig:hourly_distribution}, which plots the hourly travel time distribution with and without the D-PA implementation. Reductions in average travel times are observed at all times, predominantly during the rush hour traffic. Though using clearways leads to a notable increase in the average walking distance in Table \ref{ta:real_world}, the reduction in average time loss is not as substantial as ours. S-PA alternative achieves similar results to clearways, highlighting the necessity of our agent-based dynamic on-street parking space configuration mechanism over restricting the usage of few parking spaces at all times. The proposed Deep Q-learning architecture outperforms all other reinforcement learning based architectures, showcasing the importance of capturing spatio-temporal correlations through LSTM and GAT layers when deciding on the optimal actions.
 
\begin{figure*}[htbp]
\centering

\setlength{\tabcolsep}{0pt} 
\renewcommand{\arraystretch}{1.8} 

\begin{tabular}{>{\raggedleft}p{0.9cm} cccc} 

     &
    \includegraphics[width=0.22\textwidth]{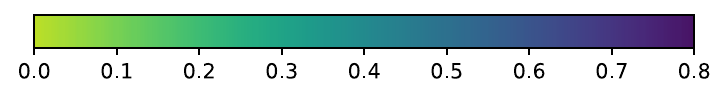} &
    \includegraphics[width=0.22\textwidth]{legend_po.pdf} &
    \includegraphics[width=0.22\textwidth]{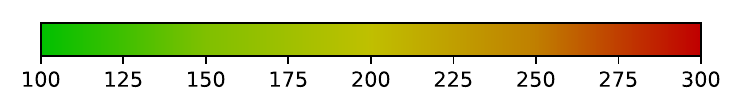} &
    \includegraphics[width=0.22\textwidth]{legend.pdf} \\
    
    \adjustbox{valign=m}{\footnotesize \makecell {8 AM - \\ 9 AM}} &
    \adjustbox{valign=m}{\includegraphics[width=0.237\textwidth]{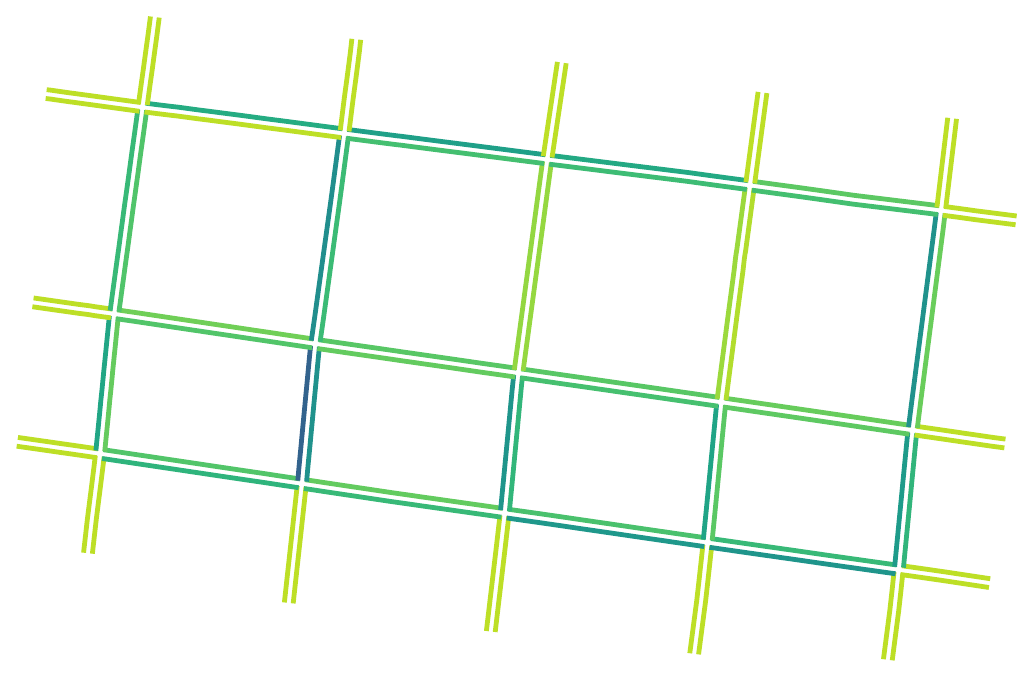}} &
    \adjustbox{valign=m}{\includegraphics[width=0.237\textwidth]{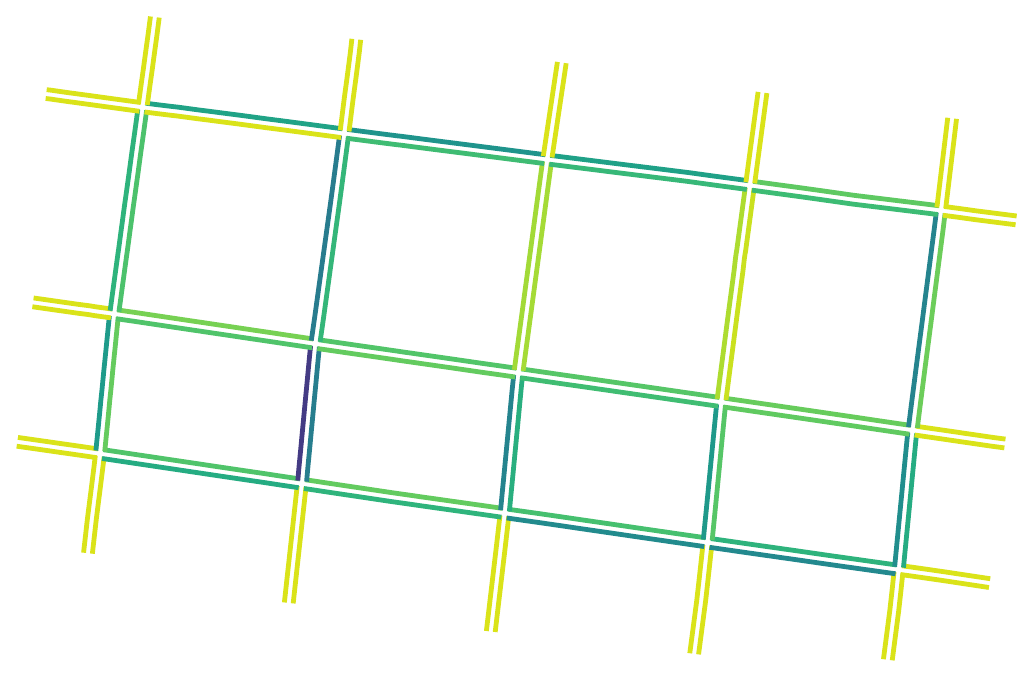}} &
    \adjustbox{valign=m}{\includegraphics[width=0.237\textwidth]{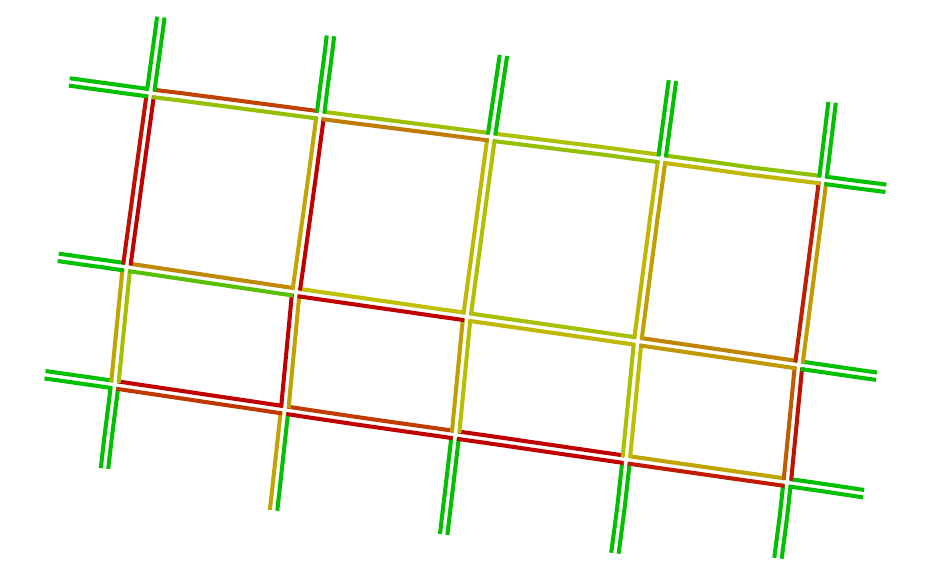}} &
    \adjustbox{valign=m}{\includegraphics[width=0.237\textwidth]{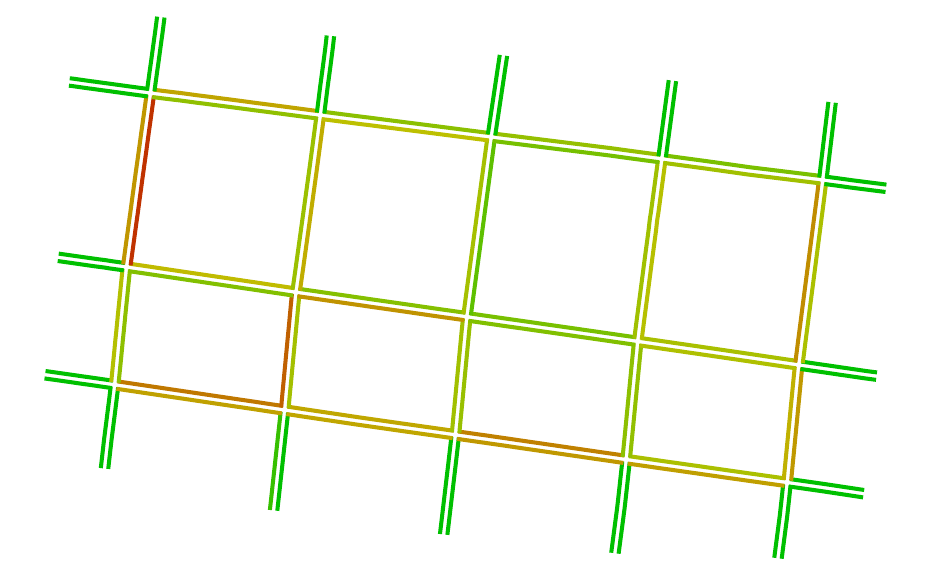}} \\
    
    \adjustbox{valign=m}{\footnotesize \makecell {11 AM - \\ 12 PM}} &
    \adjustbox{valign=m}{\includegraphics[width=0.237\textwidth]{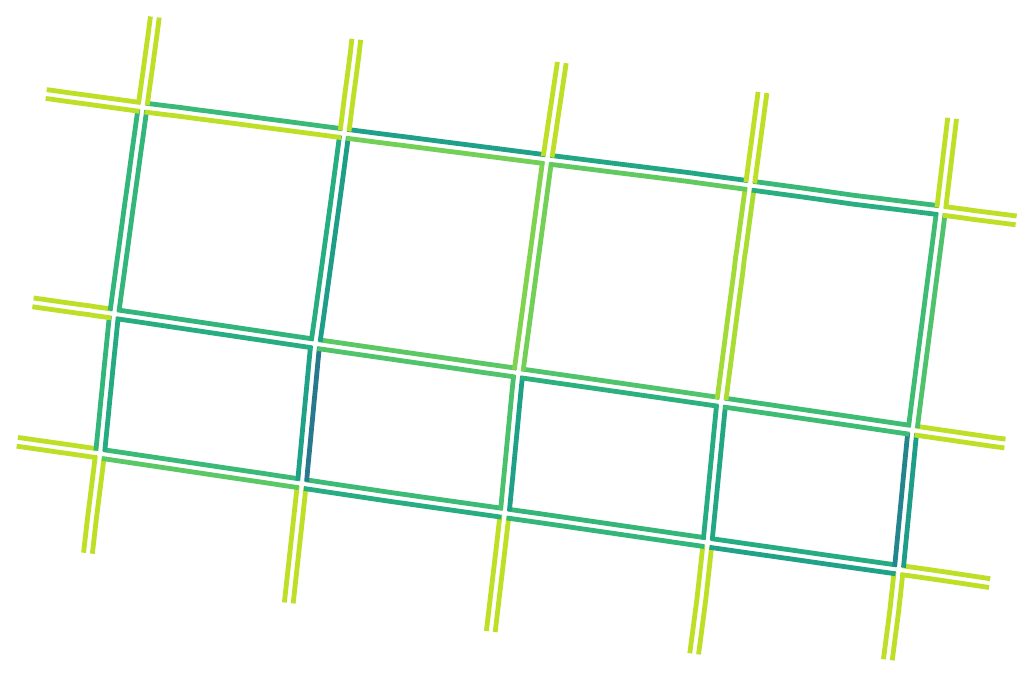}} &
    \adjustbox{valign=m}{\includegraphics[width=0.237\textwidth]{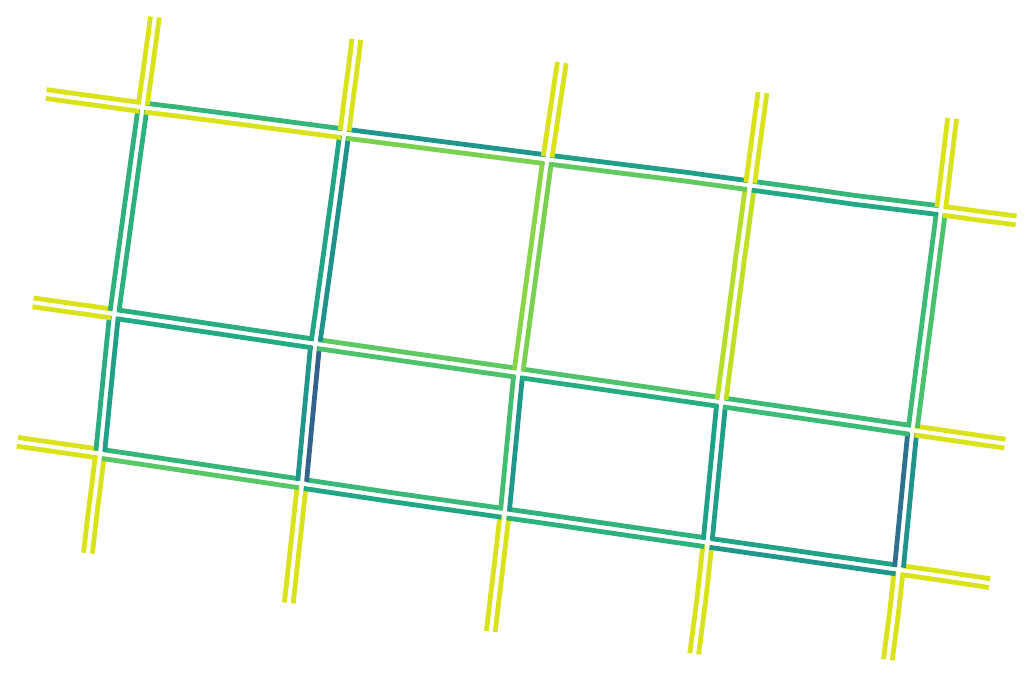}} &
    \adjustbox{valign=m}{\includegraphics[width=0.237\textwidth]{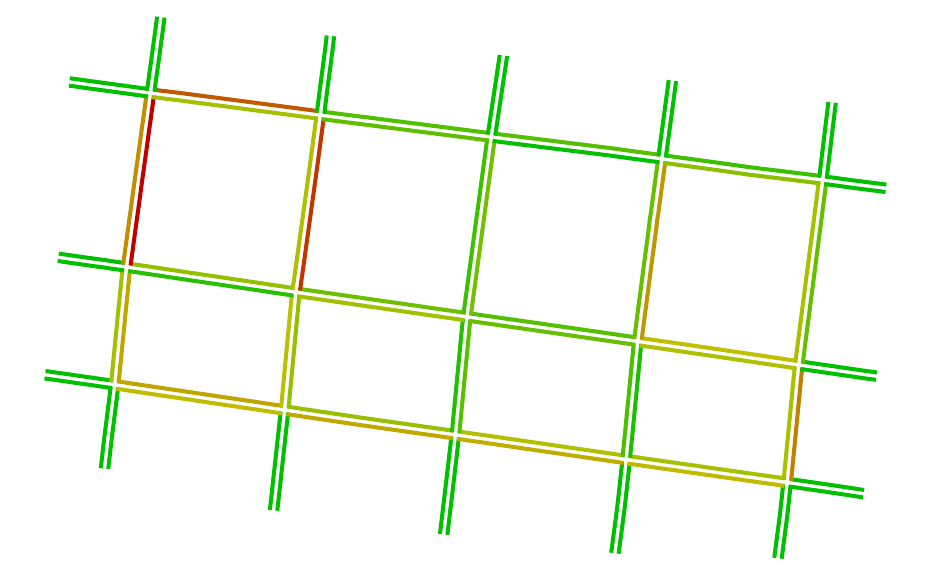}} &
    \adjustbox{valign=m}{\includegraphics[width=0.237\textwidth]{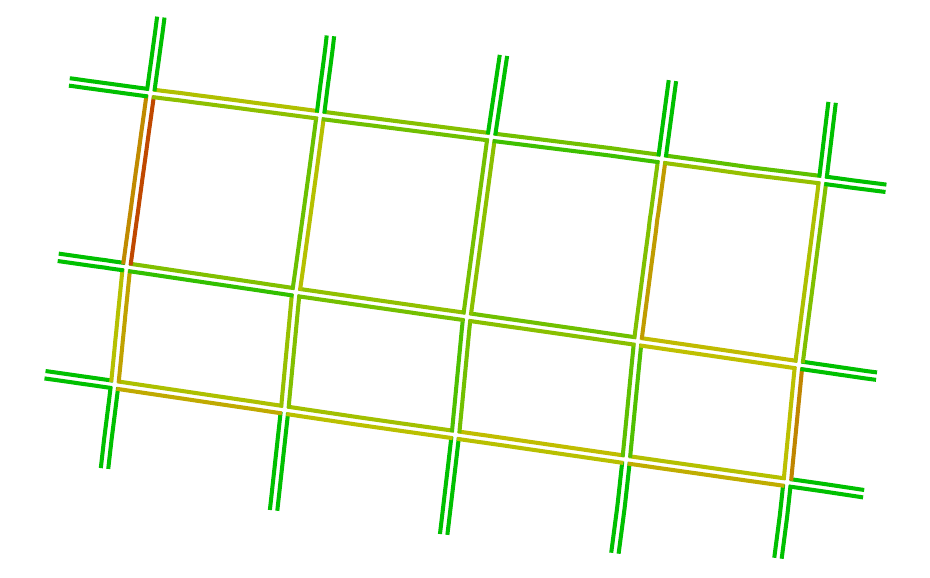}} \\
    
    \adjustbox{valign=m}{\footnotesize \makecell {2 PM - \\ 3 PM}} &
    \adjustbox{valign=m}{\includegraphics[width=0.237\textwidth]{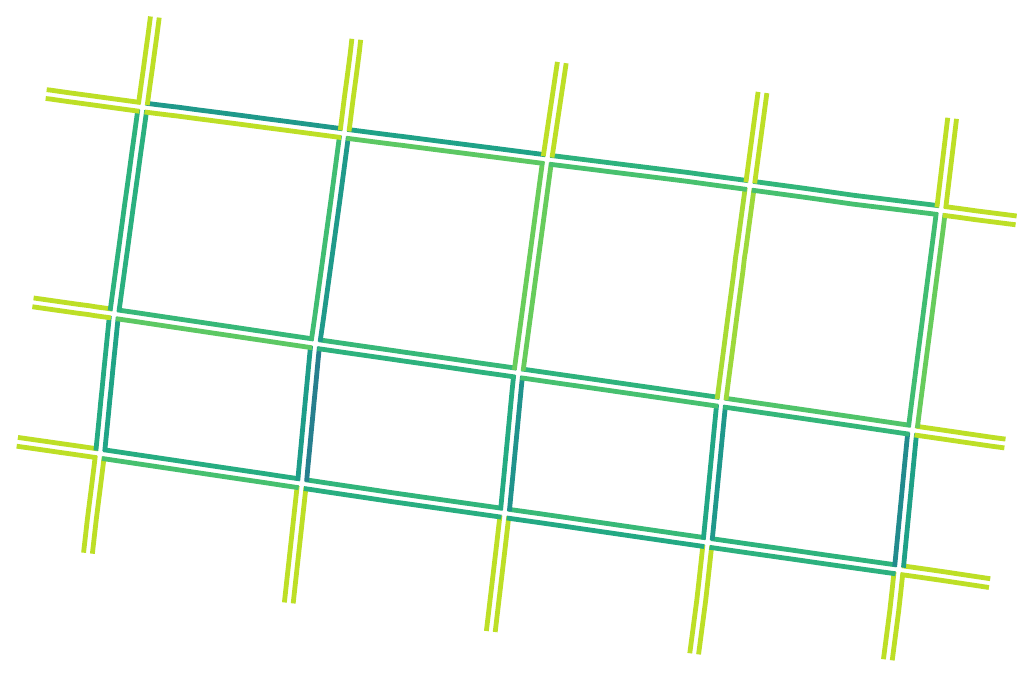}} &
    \adjustbox{valign=m}{\includegraphics[width=0.237\textwidth]{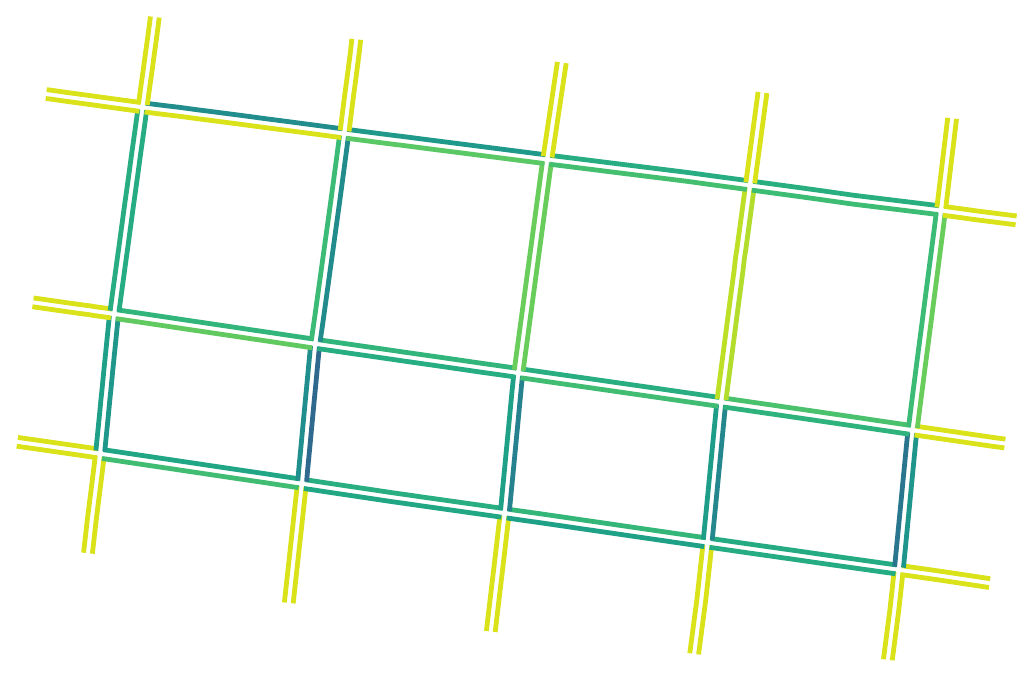}} &
    \adjustbox{valign=m}{\includegraphics[width=0.237\textwidth]{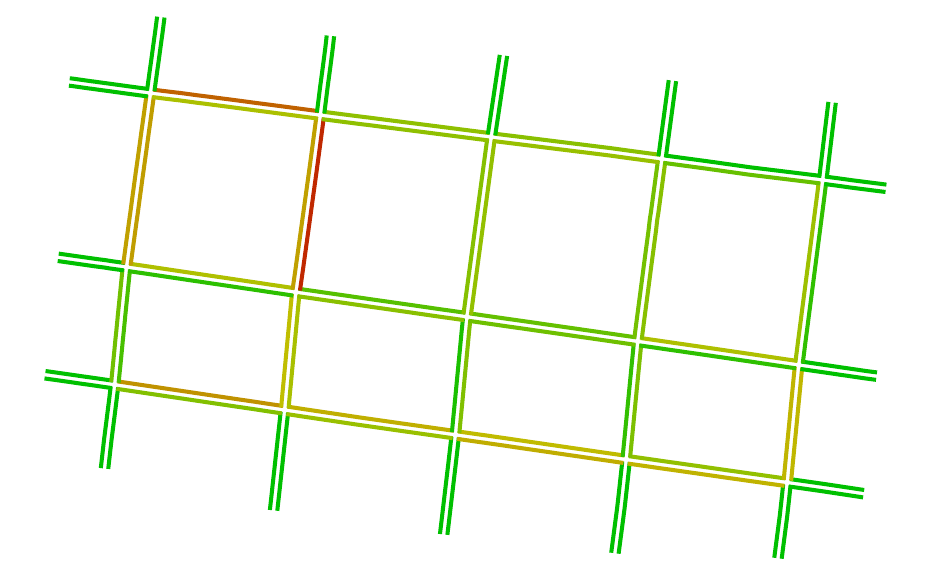}} &
    \adjustbox{valign=m}{\includegraphics[width=0.237\textwidth]{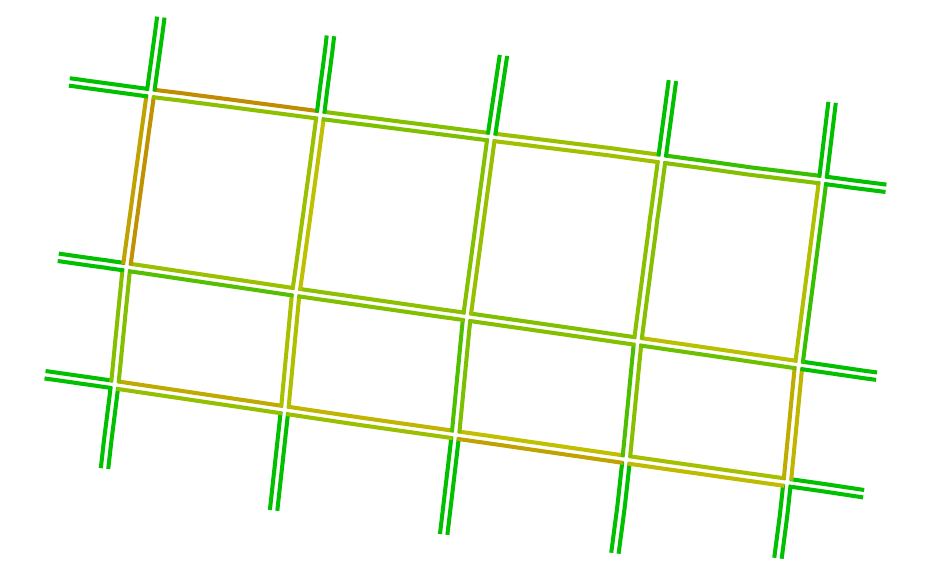}} \\

   \adjustbox{valign=m}{\footnotesize \makecell {5 PM - \\ 6 PM}} &
    \adjustbox{valign=m}{\includegraphics[width=0.237\textwidth]{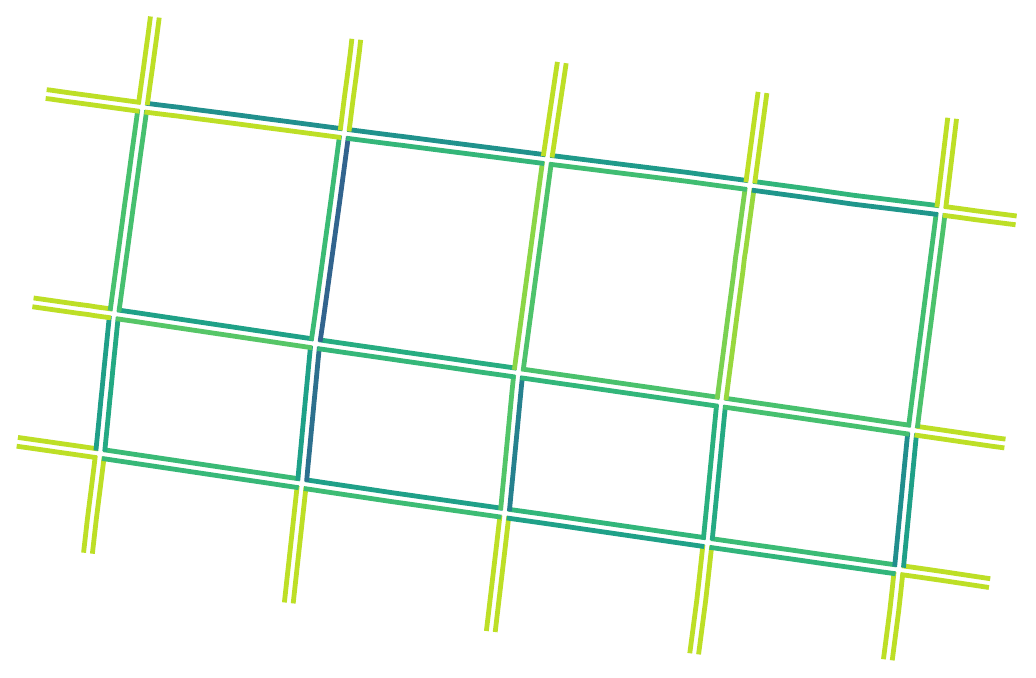}} &
    \adjustbox{valign=m}{\includegraphics[width=0.237\textwidth]{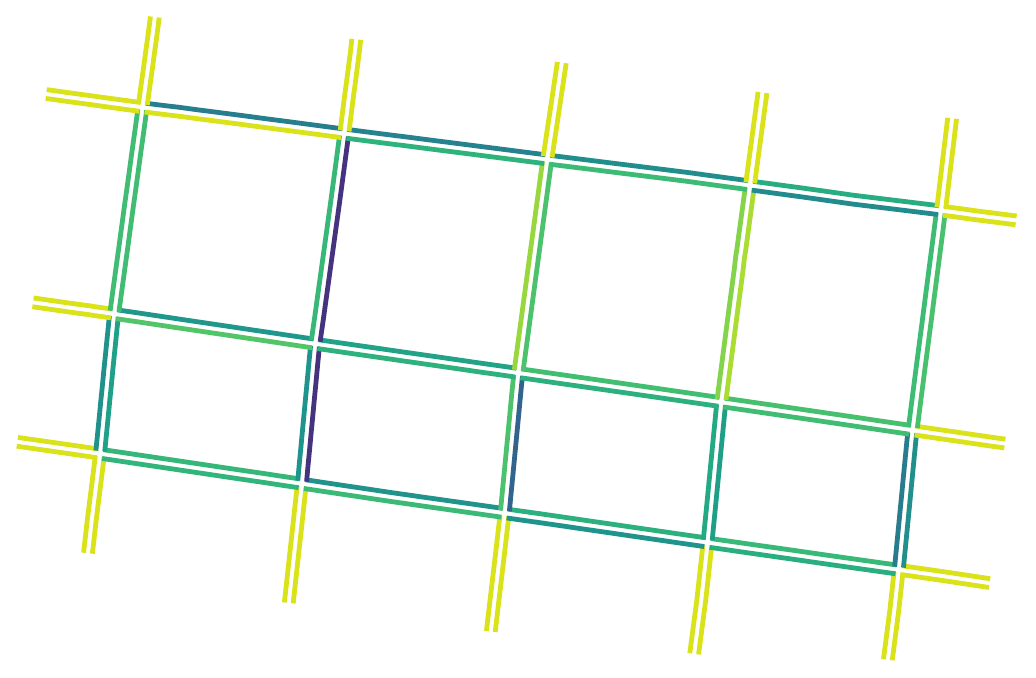}} &
    \adjustbox{valign=m}{\includegraphics[width=0.237\textwidth]{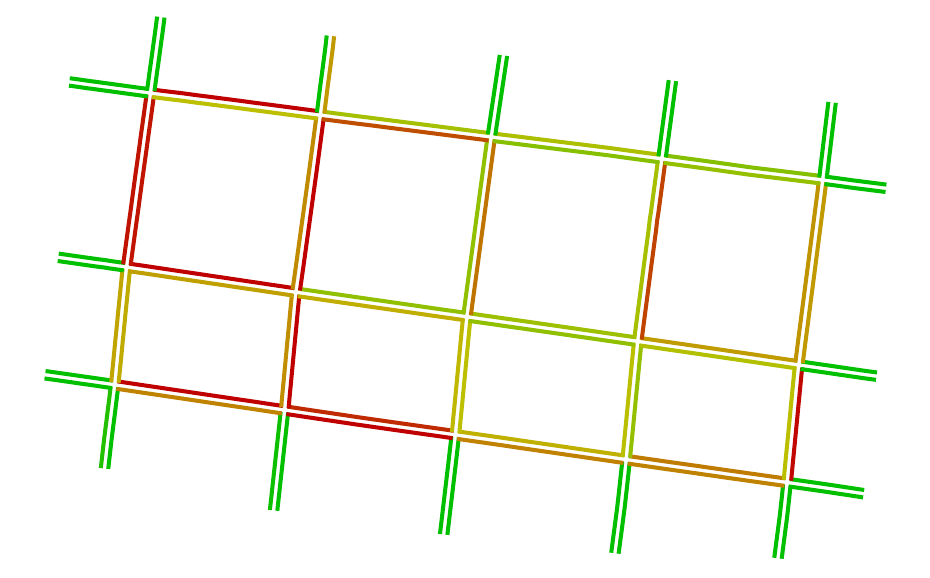}} &
    \adjustbox{valign=m}{\includegraphics[width=0.237\textwidth]{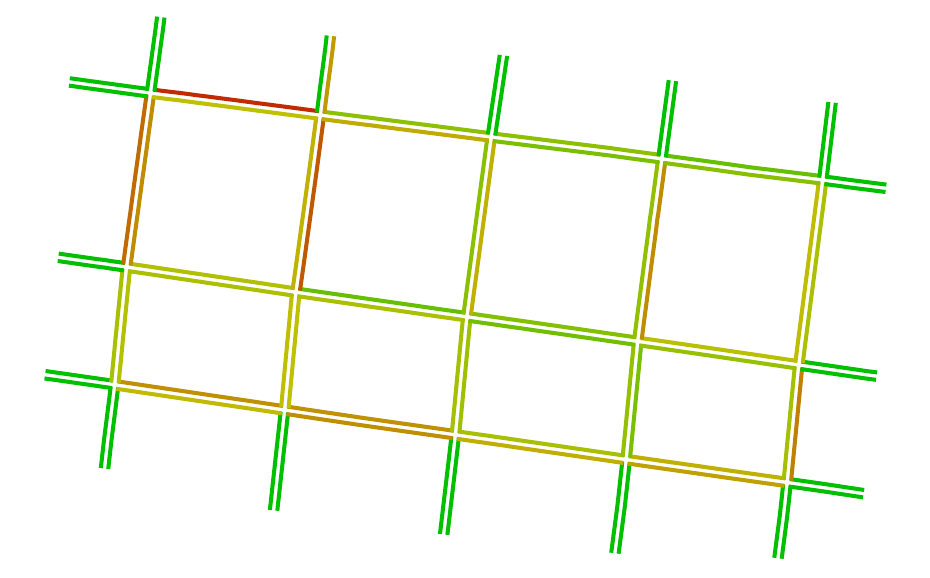}} \\
    
    & \small Parking occupancy (No-PA) & \small Parking occupancy (D-PA) & \small Average travel time (s) (No-PA) & \small Average travel time (s) (D-PA)
\end{tabular}

\caption{Parking occupancy and average travel time variation, with and without D-PA, at different times of the day. While parking occupancy has increased slightly with D-PA, average travel time could be reduced at all times, notably during rush hours.}
\label{fi:graphical}
\end{figure*}

\subsubsection{Emissions}

Figure \ref{fi:emissions} illustrates the percentages of reduction in gas emissions under the proposed D-PA mechanism, compared with No-PA. The analysis consists of carbon dioxide (CO2), carbon monoxide (CO), hydrocarbons (HC), particular materials (PMx) and nitrogen oxides (NOx) emissions associated with the experiments conducted on real-world data. It could be observed that significant reductions could be obtained reaching upto 40\%, which highlights the positive environmental impact of the proposed solution. 

\subsubsection{Temporal Heatmaps}

Figure \ref{fi:graphical} shows the parking occupancy and the average travel time variation in each lane using heatmaps, with and without D-PA, for different time periods of the day. The parking occupancy has been slightly increased in most cases, since few parking spaces will be restricted from using, under D-PA. The average travel time has been reduced in all time periods by incorporating D-PA, significantly during morning and evening rush hours. 

 \subsection{Sensitivity Analysis}

  \begin{table}[t]
\centering
\small
\begin{tblr}{
  width = \linewidth,
  colspec = {Q[m,208]Q[m,208]Q[m,208]Q[m,258]},
  cells = {c},
  hline{1,2,6},
  vline{2,3,4},
}
Parking Probability & DQN & DQN + LSTM & DQN + LSTM + GAT \\
0.10 & 7.46\% & 11.52\% & 13.03\%  \\
0.20 & 16.60\% & 21.25\% & 25.58\% \\
0.30 & 27.47\% & 30.31\% & 39.70\% \\
0.40 & 32.93\% & 38.52\% & 47.79\% 
\end{tblr}
\vspace{1em}
\caption{Ablation study on proposed lane level network architecture on real-world data. By incorporating spatio-temporal correlations through LSTM and GAT networks, the average time loss reduction percentages have been further increased..}
\label{ta:ablation_study}
\end{table}

 The experimental results of the sensitivity analysis conducted using synthetic data are shown in Figure \ref{fi:sensitivity_analysis}. As shown in Figure 7a, our method results in higher gains in travel time loss, when the roads are more congested. As shown in Figure 7b and Figure 7c, the travel time loss has increased exponentially at a much higher rate under C-PA, with increasing parking probability and increasing parking duration. This is because, there is no clear direction of traffic in a random traffic scenario, and with increased need of parking, more vehicles would have to cruise around in a clearway setup. However, our proposed two-layer multi agent based architecture consistently provides robust performance, even with increasing parking probability and parking duration. As shown in Figure 7d, the average travel time loss generally increases with the grid size, due to the increase in the route lengths of the vehicles. Nevertheless, D-PA consistently achieves the lowest average travel time loss, outperforming all other alternatives, across the considered networks with varying grid sizes. 

 \subsection{Ablation Study}

We conduct an ablation study to better understand the contribution of each abstraction level of the proposed methodology. We evaluate the performance of the RL model used as the lane level agent on real-world data, comparing our architecture (\textit{DQN + LSTM + GAT}) with a vanilla DQN model (\textit{DQN}), and a DQN model capturing only temporal behaviour through LSTM layers (\textit{DQN + LSTM}). The results are presented in Table \ref{ta:ablation_study}, which provide evidence that the average time loss could be significantly reduced by incorporating traffic congestion variation of the selected lane and spatio-temporal behaviour of traffic congestion of adjacent lane segments through LSTM and GAT layers. 

\section{Discussion on Real-World Deployment}

Though, traffic rules and regulations have been imposed mostly using static signs currently, the developments in vehicle-to-infrastructure connectivity (V2I) enable the application of dynamic and sophisticated solutions for achieving more efficient and sustainable transportation systems. For instance, traffic signal control solutions that anticipate the approaching traffic flows and optimize the phases dynamically lead to significant reductions in travel time and emissions when compared to actuated or fixed-time control \cite{chumarl}. Within the context of parking, map based parking availability applications such as \cite{com_opendata} have already been developed which provide information on real-time parking availability for the users. The D-PA architecture can be deployed similarly using a mobile application which provides parking restriction information for users in real-time. Similar to electronic signs that display dynamic speed limits \cite{vsl}, electronic displays can also be used to indicate whether a parking space can be used or not. However, the latter solution would result in a higher infrastructure cost when compared with using mobile phones or in-car display units. Furthermore, with the evolution of V2I technologies, the vehicles themselves can obtain real-time parking restriction information. The highly scalable nature of the two-layered D-PA architecture makes it well suitable for deploying in large road networks with multiple road segments. However, compliance with regulatory frameworks and social acceptance can be identified as key challenges associated with real-world implementation.

\section{Conclusion}
In this work, we formalized dynamic configuration of on-street parking spaces as an optimization problem, and proposed a multi agent reinforcement learning based two-layer solution architecture for configuring on-street parking spaces dynamically and efficiently, based on the observed traffic conditions. While the bottom layer RL agents focus on deciding the number of parking spaces to be cleared in each lane, the higher level block agents control the actions of the lower level agents, and maintain a sufficient level of parking availability around the neighbourhood. A novel Deep Q-learning network architecture was introduced to train the lane level agents, which incorporates LSTM and GAT networks to capture spatio-temporal correlations, and provide a thorough understanding of the environment, which would help the agent to take optimal actions that align with future traffic conditions. The methodology was validated using both synthetic and real-world data using the SUMO simulation platform, including a sensitivity analysis and an ablation study. The proposed system performs remarkably well, achieving high percentage reductions in average travel time loss of vehicles, reaching upto 47\%, at a less than two meter increase in walking distance. As future work, we aim to extend our solution framework to autonomous and mixed traffic scenarios, in which the optimization problem becomes further complicated. We believe that our work is a promising step towards reducing traffic congestion associated with parking, which has been a growing concern over the last few decades.

\bibliographystyle{IEEEtran}
\bibliography{references}

\begin{thebibliography}{10}
\providecommand{\url}[1]{#1}
\csname url@samestyle\endcsname
\providecommand{\newblock}{\relax}
\providecommand{\bibinfo}[2]{#2}
\providecommand{\BIBentrySTDinterwordspacing}{\spaceskip=0pt\relax}
\providecommand{\BIBentryALTinterwordstretchfactor}{4}
\providecommand{\BIBentryALTinterwordspacing}{\spaceskip=\fontdimen2\font plus
\BIBentryALTinterwordstretchfactor\fontdimen3\font minus \fontdimen4\font\relax}
\providecommand{\BIBforeignlanguage}[2]{{%
\expandafter\ifx\csname l@#1\endcsname\relax
\typeout{** WARNING: IEEEtran.bst: No hyphenation pattern has been}%
\typeout{** loaded for the language `#1'. Using the pattern for}%
\typeout{** the default language instead.}%
\else
\language=\csname l@#1\endcsname
\fi
#2}}
\providecommand{\BIBdecl}{\relax}
\BIBdecl

\bibitem{prabu}
P.~K.J., H.~K. A.N, and S.~Bhatnagar, ``Multi-agent reinforcement learning for traffic signal control,'' in \emph{17th IEEE International Conference on Intelligent Transportation Systems (ITSC)}, 2014, pp. 2529--2534.

\bibitem{chumarl}
T.~Chu, J.~Wang, L.~Codecà, and Z.~Li, ``Multi-agent deep reinforcement learning for large-scale traffic signal control,'' \emph{IEEE Transactions on Intelligent Transportation Systems}, vol.~21, no.~3, pp. 1086--1095, 2020.

\bibitem{wumarl}
T.~Wu, P.~Zhou, K.~Liu, Y.~Yuan, X.~Wang, H.~Huang, and D.~O. Wu, ``Multi-agent deep reinforcement learning for urban traffic light control in vehicular networks,'' \emph{IEEE Transactions on Vehicular Technology}, vol.~69, no.~8, pp. 8243--8256, 2020.

\bibitem{WANGmarl}
T.~Wang, J.~Cao, and A.~Hussain, ``Adaptive traffic signal control for large-scale scenario with cooperative group-based multi-agent reinforcement learning,'' \emph{Transportation Research Part C: Emerging Technologies}, vol. 125, pp. 1--27, 2021.

\bibitem{mamarl}
J.~Ma and F.~Wu, ``Feudal multi-agent deep reinforcement learning for traffic signal control,'' in \emph{19th International Conference on Autonomous Agents and Multiagent Systems}, 2020, pp. 816--824.

\bibitem{milp}
Y.~Geng and C.~G. Cassandras, ``New “smart parking” system based on resource allocation and reservations,'' \emph{IEEE Transactions on Intelligent Transportation Systems}, vol.~14, no.~3, pp. 1129--1139, 2013.

\bibitem{taiwan}
E.~H.-K. Wu, J.~Sahoo, C.-Y. Liu, M.-H. Jin, and S.-H. Lin, ``Agile urban parking recommendation service for intelligent vehicular guiding system,'' \emph{IEEE Intelligent Transportation Systems Magazine}, vol.~6, no.~1, pp. 35--49, 2014.

\bibitem{secon}
K.~S. Liu, J.~Gao, X.~Wu, and S.~Lin, ``On-street parking guidance with real-time sensing data for smart cities,'' in \emph{15th Annual IEEE International Conference on Sensing, Communication, and Networking (SECON)}, 2018, pp. 1--9.

\bibitem{sun2024}
H.~Sun, X.~Huang, and W.~Ma, ``Beyond prediction: On-street parking recommendation using heterogeneous graph-based list-wise ranking,'' \emph{IEEE Transactions on Intelligent Transportation Systems}, vol.~25, no.~6, pp. 5892--5903, 2024.

\bibitem{markov}
J.~Xiao and Y.~Lou, ``An online reinforcement learning approach for user-optimal parking searching strategy exploiting unique problem property and network topology,'' \emph{IEEE Transactions on Intelligent Transportation Systems}, vol.~23, no.~7, pp. 8157--8169, 2022.

\bibitem{com_opendata}
Datavic, ``On-street parking bay sensors,'' \url{https://data.melbourne.vic.gov.au/explore/dataset/on-street-parking-bay-sensors/custom/}.

\bibitem{zhang2011micro}
K.~Zhang and A.~Excell, ``A micro-simulation approach to quantifying clearway benefits,'' in \emph{Australasian Transport Research Forum}, 2011, pp. 1--11.

\bibitem{SUMO2018}
P.~A. Lopez, M.~Behrisch, L.~Bieker-Walz, J.~Erdmann, Y.-P. Fl{\"o}tter{\"o}d, R.~Hilbrich, L.~L{\"u}cken, J.~Rummel, P.~Wagner, and E.~Wie{\ss}ner, ``Microscopic traffic simulation using sumo,'' in \emph{21st IEEE International Conference on Intelligent Transportation Systems}, 2018, pp. 2575--2582.

\bibitem{DQN}
V.~Mnih, K.~Kavukcuoglu, D.~Silver, A.~Graves, I.~Antonoglou, D.~Wierstra, and M.~A. Riedmiller, ``Playing atari with deep reinforcement learning,'' \emph{arXiv preprint arXiv:1312.5602}, 2013.

\bibitem{LSTM}
S.~Hochreiter and J.~Schmidhuber, ``Long short-term memory,'' \emph{Neural Computation}, vol.~9, no.~8, p. 1735–1780, 1997.

\bibitem{GAT}
P.~Veli{\v{c}}kovi{\'c}, G.~Cucurull, A.~Casanova, A.~Romero, P.~Li{\`o}, and Y.~Bengio, ``Graph attention networks,'' in \emph{6th International Conference on Learning Representations (ICLR)}, 2018, p. 1–12.

\bibitem{guohong}
H.~Guo, W.~Wang, and W.~Guo, ``Micro-simulation study on the effect of on-street parking on vehicular flow,'' in \emph{15th IEEE International Conference on Intelligent Transportation Systems}, 2012, pp. 1840--1845.

\bibitem{sugiarto}
S.~Sugiarto and T.~Limanoond, ``Impact of on-street parking on urban arterial performance: A quantitative study on travel speed and capacity deterioration,'' \emph{Aceh International Journal of Science and Technology}, vol.~2, no.~2, pp. 63--69, 2013.

\bibitem{caojin}
J.~Cao, M.~Menendez, and V.~Nikias, ``The effects of on-street parking on the service rate of nearby intersections,'' \emph{Journal of Advanced Transportation}, vol.~50, p. 406–420, 2016.

\bibitem{genetic}
Q.~Ye, S.~M. Stebbins, Y.~Feng, E.~Candela, M.~Stettler, and P.~Angeloudis, ``Intelligent management of on-street parking provision for the autonomous vehicles era,'' in \emph{23rd IEEE International Conference on Intelligent Transportation Systems (ITSC)}, 2020, pp. 1--7.

\bibitem{A2C}
V.~Mnih, A.~P. Badia, M.~Mirza, A.~Graves, T.~Harley, T.~P. Lillicrap, D.~Silver, and K.~Kavukcuoglu, ``Asynchronous methods for deep reinforcement learning,'' in \emph{Proceedings of the 33rd International Conference on Machine Learning}, 2016, pp. 1928--1937.

\bibitem{yincurbside}
S.~Yin, Z.~Cui, and Y.~Wang, ``Reinforcement learning for curbside space management with infrastructure autonomy and mixed vehicle connectivity,'' in \emph{25th IEEE International Conference on Intelligent Transportation Systems (ITSC)}, 2022, pp. 3276--3282.

\bibitem{nazirzoning}
N.~Nazir, C.~Dowling, S.~Choudhury, S.~Zoepf, and K.~Ma, ``Optimal, centralized dynamic curbside parking space zoning,'' in \emph{25th IEEE International Conference on Intelligent Transportation Systems (ITSC)}, 2022, pp. 91--98.

\bibitem{SHOUP2006479}
D.~C. Shoup, ``Cruising for parking,'' \emph{Transport Policy}, vol.~13, no.~6, pp. 479--486, 2006.

\bibitem{basu}
P.~Basu and T.~Little, ``Networked parking spaces: architecture and applications,'' in \emph{56th IEEE Vehicular Technology Conference}, 2002, pp. 1153--1157.

\bibitem{vissim2}
M.~Fellendorf and P.~Vortisch, ``{Microscopic Traffic Flow Simulator VISSIM},'' in \emph{{Fundamentals of Traffic Simulation}}, 2010, vol. 145, pp. 63--93.

\bibitem{gunarathna2021real}
U.~Gunarathna, H.~Xie, E.~Tanin, S.~Karunasekara, and R.~Borovica-Gajic, ``Real-time lane configuration with coordinated reinforcement learning,'' in \emph{The European Conference on Machine Learning and Principles and Practice of Knowledge Discovery in Databases (ECML PKDD)}, 2021, pp. 291--307.

\bibitem{marlparking}
X.~Zhang, C.~Zhao, F.~Liao, X.~Li, and Y.~Du, ``Online parking assignment in an environment of partially connected vehicles: A multi-agent deep reinforcement learning approach,'' \emph{Transportation Research Part C: Emerging Technologies}, vol. 138, pp. 1--22, 2022.

\bibitem{qmix}
T.~Rashid, M.~Samvelyan, C.~Schroeder, G.~Farquhar, J.~Foerster, and S.~Whiteson, ``{QMIX}: Monotonic value function factorisation for deep multi-agent reinforcement learning,'' in \emph{Proceedings of the 35th International Conference on Machine Learning}, 2018, pp. 4295--4304.

\bibitem{scats}
Datavic, ``Traffic signal volume data,'' \url{https://discover.data.vic.gov.au/dataset/traffic-signal-volume-data}.

\bibitem{datasheets}
------, ``Traffic signal configuration data sheets,'' \url{https://discover.data.vic.gov.au/dataset/traffic-signal-configuration-data-sheets}.

\bibitem{parking_data}
------, ``On street car parking sensor data 2018,'' \url{https://discover.data.vic.gov.au/dataset/on-street-car-parking-sensor-data-2018}.

\bibitem{pytorch}
A.~Paszke, S.~Gross, F.~Massa, A.~Lerer, J.~Bradbury, G.~Chanan, T.~Killeen, Z.~Lin, N.~Gimelshein, L.~Antiga, A.~Desmaison, A.~Kopf, E.~Yang, Z.~DeVito, M.~Raison, A.~Tejani, S.~Chilamkurthy, B.~Steiner, L.~Fang, J.~Bai, and S.~Chintala, ``Pytorch: An imperative style, high-performance deep learning library,'' in \emph{Proceedings of the 33rd International Conference on Neural Information Processing Systems}, 2019, pp. 8026--8037.

\bibitem{clearway_data}
Datavic, ``Clearways data,'' \url{https://vicroadsopendata-vicroadsmaps.opendata.arcgis.com/datasets/53aabcfbf96e4b4bbd288534e053dee5/explore}.

\bibitem{ppo}
J.~Schulman, F.~Wolski, P.~Dhariwal, A.~Radford, and O.~Klimov, ``Proximal policy optimization algorithms,'' \emph{arXiv preprint arXiv:1707.06347}, 2018.

\bibitem{ddqn}
H.~V. Hasselt, A.~Guez, and D.~Silver, ``Deep reinforcement learning with double q-learning,'' in \emph{Proceedings of the 30th AAAI Conference on Artificial Intelligence}, 2016, p. 2094–2100.

\bibitem{dudqn}
Z.~Wang, T.~Schaul, M.~Hessel, H.~Van~Hasselt, M.~Lanctot, and N.~De~Freitas, ``Dueling network architectures for deep reinforcement learning,'' in \emph{Proceedings of the 33rd International Conference on International Conference on Machine Learning}, 2016, p. 1995–2003.

\bibitem{vsl}
M.~Papageorgiou, E.~Kosmatopoulos, and I.~Papamichail, ``Effects of variable speed limits on motorway traffic flow,'' \emph{Transportation Research Record}, vol. 2047, no.~1, pp. 37--48, 2008.

\end{thebibliography}

\end{document}